\documentclass[10pt,british,runningheads]{llncs}
\usepackage[T1]{fontenc}
\usepackage[latin9]{inputenc}
\usepackage{color}
\usepackage{babel}
\usepackage{amsmath}
\usepackage{graphicx}
\PassOptionsToPackage{normalem}{ulem}
\usepackage{ulem}
\usepackage[unicode=true,pdfusetitle,
 bookmarks=true,bookmarksnumbered=false,bookmarksopen=false,
 breaklinks=false,pdfborder={0 0 1},backref=false,colorlinks=false]
 {hyperref}

\makeatletter

\providecommand{\tabularnewline}{\\}
\newcommand{\lyxdot}{.}

\usepackage{graphicx}
\usepackage{color}
\usepackage{colortbl}
\usepackage[width=122mm,left=12mm,paperwidth=146mm,height=193mm,top=12mm,paperheight=217mm]{geometry}
\usepackage{rotating}
\usepackage[font=small]{subfig}
\definecolor{lightgray}{gray}{0.8}
\usepackage{pdflscape}

\@ifundefined{showcaptionsetup}{}{%
 \PassOptionsToPackage{caption=false}{subfig}}
\usepackage{subfig}
\makeatother

\begin{document}

\title{Improved Image Boundaries for Better Video Segmentation}

\author{Anna Khoreva\textsuperscript{1}\hspace{1.5em}Rodrigo Benenson\textsuperscript{1}\hspace{1.5em}Fabio
Galasso\textsuperscript{2}\hspace{1.5em}Matthias Hein\textsuperscript{3}\hspace{1.5em}Bernt
Schiele\textsuperscript{1}}

\authorrunning{A. Khoreva, R. Benenson, F. Galasso, M. Hein and B. Schiele}

\institute{\textsuperscript{1}Max Planck Institute for Informatics, Saarbr{\"u}cken,
Germany\\ \textsuperscript{2}OSRAM Corporate Technology, Germany\\
\textsuperscript{3}Saarland University, Saarbr{\"u}cken, Germany}

\maketitle
\makeatletter 
\renewcommand{\paragraph}{%
\@startsection{paragraph}{4}%
{\z@}{1.0ex \@plus 1ex \@minus .2ex}{-1em}%
{\normalfont \normalsize \bfseries}%
}
\makeatother\setlength{\textfloatsep}{1em}\vspace{-1em}

\begin{abstract}
Graph-based video segmentation methods rely on superpixels as starting
point. While most previous work has focused on the construction of
the graph edges and weights as well as solving the graph partitioning
problem, this paper focus\textcolor{black}{es }on better superpixels
for video segmentation. We demonstrate by a comparative analysis that
superpixels extracted from boundaries perform best, and show that
boundary estimation can be significantly improved via image and time
domain cues. With superpixels generated from our better boundaries
we observe consistent improvement for two video segmentation methods
in two different datasets. 
\begin{figure*}
\begin{centering}
\vspace{-1em}
\hspace*{\fill}%
\begin{tabular}{ccccccc}
\includegraphics[bb=115bp 20bp 1030bp 648bp,clip,width=0.23\columnwidth,height=0.08\textheight]{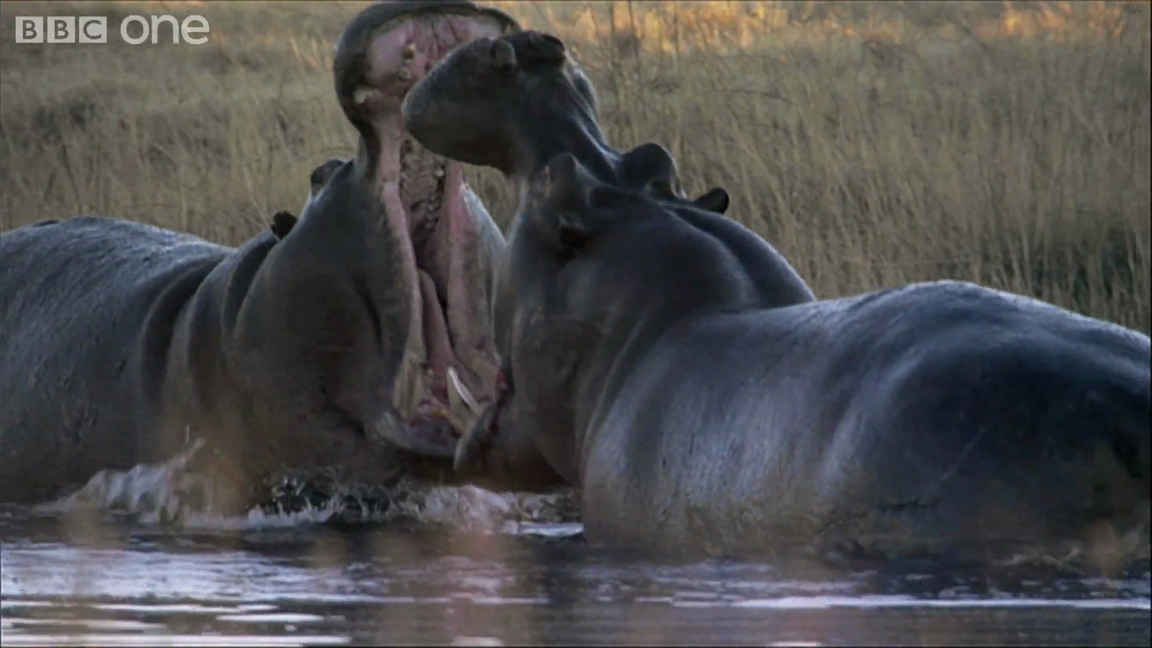} & \, & \includegraphics[bb=0bp 200bp 612bp 620bp,clip,width=0.23\columnwidth,height=0.08\textheight]{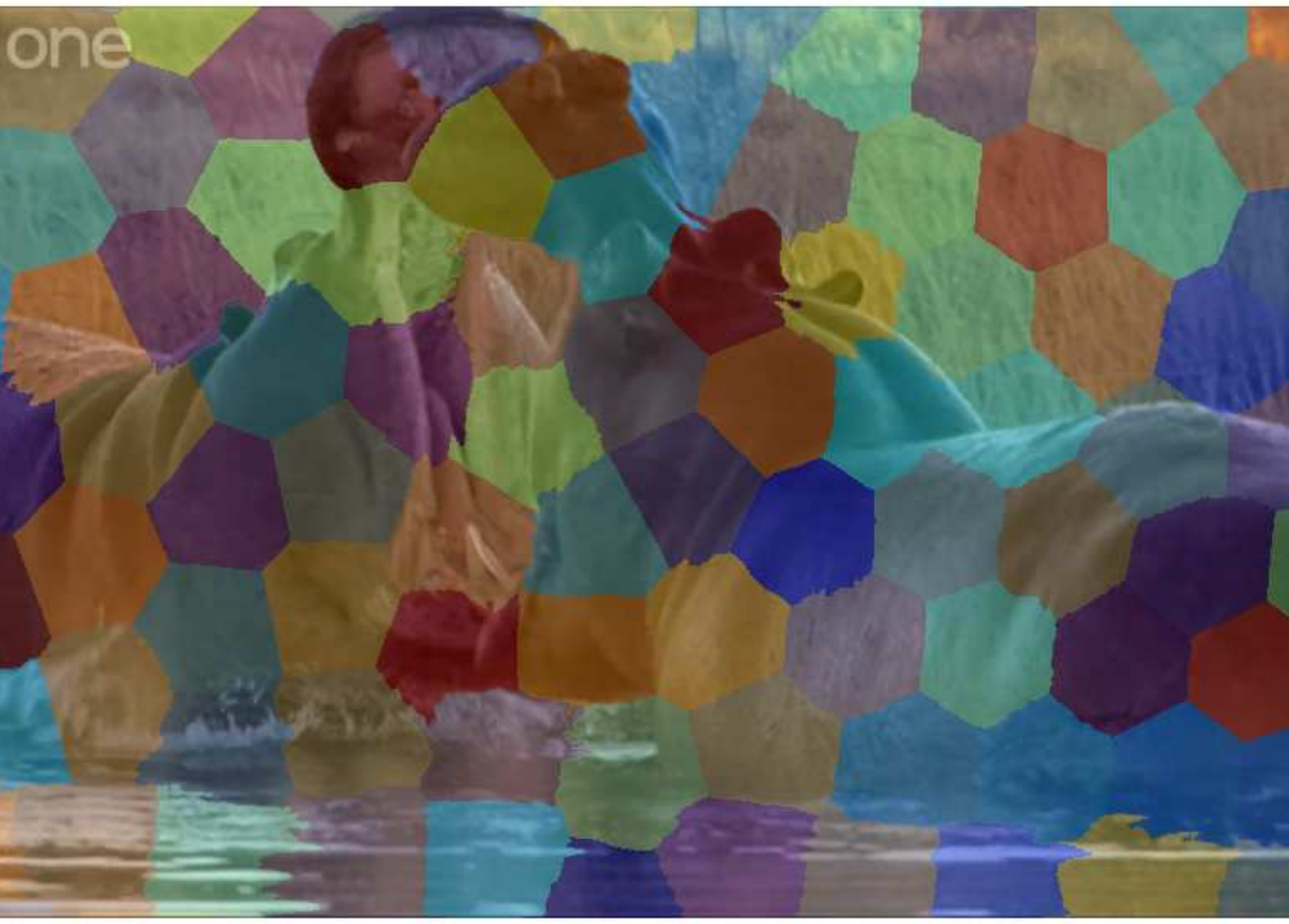} & \, & \includegraphics[bb=0bp 200bp 612bp 620bp,clip,width=0.23\columnwidth,height=0.08\textheight]{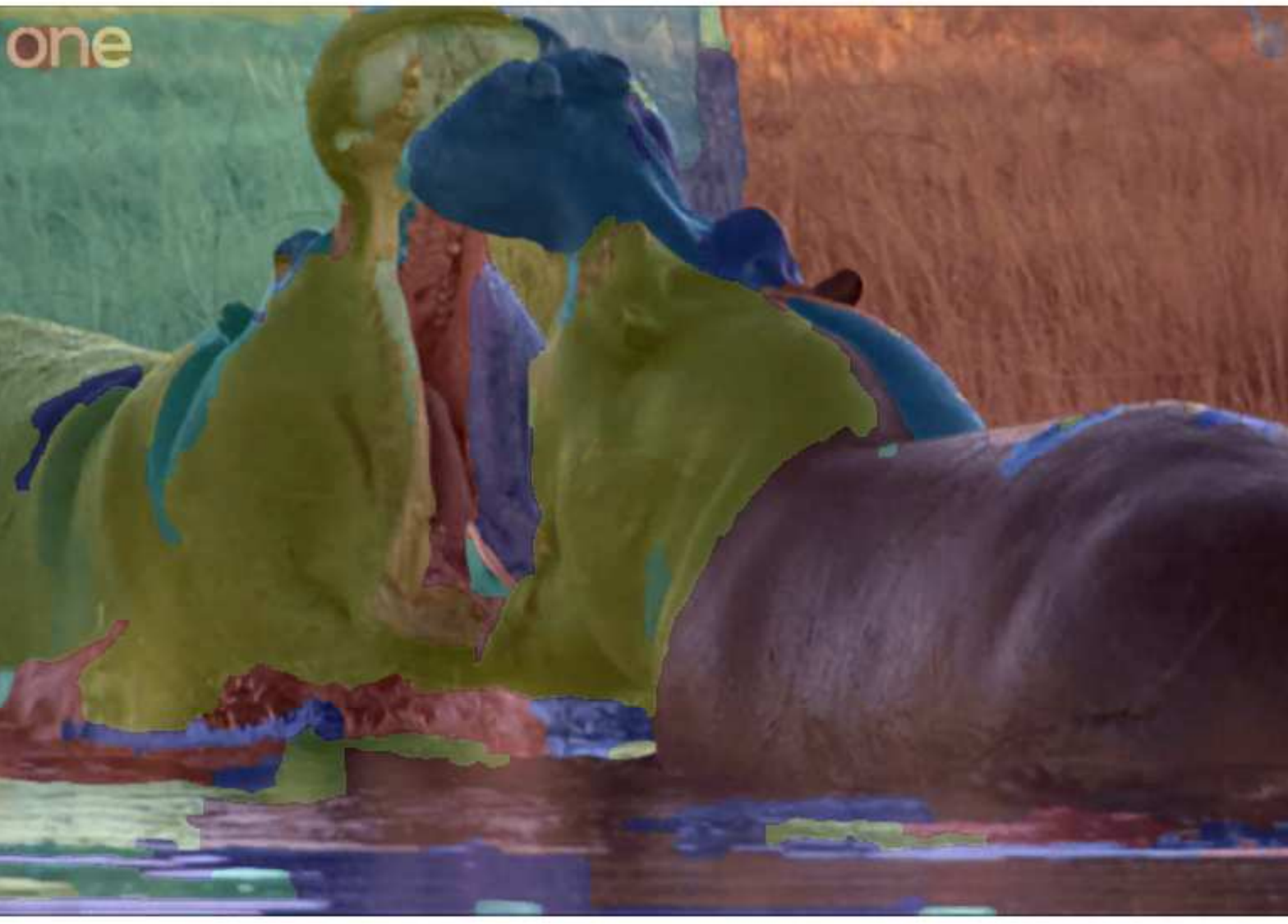} & \, & \includegraphics[bb=0bp 200bp 612bp 620bp,clip,width=0.23\columnwidth,height=0.08\textheight]{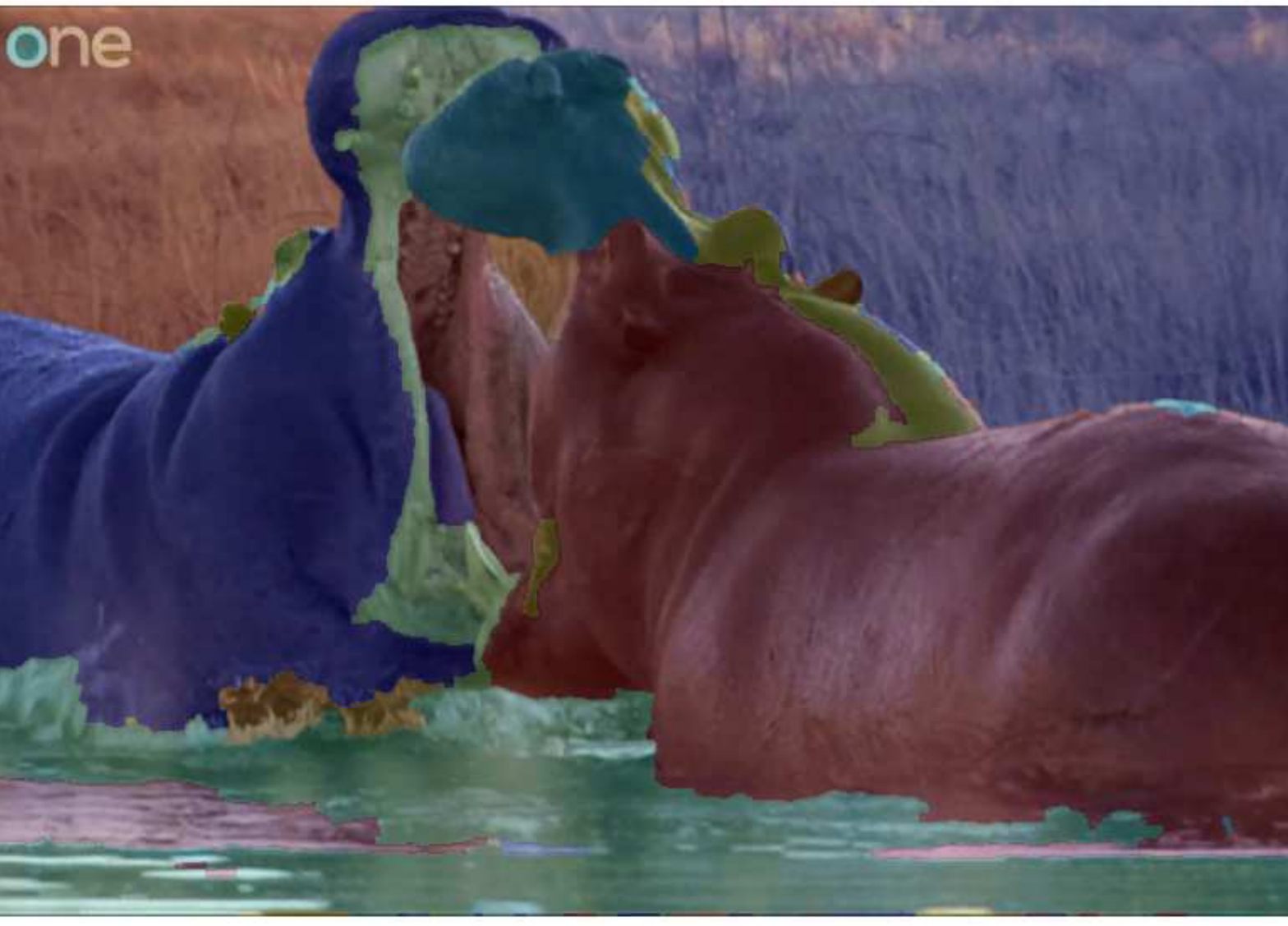}\tabularnewline
Video &  & TSP superpixels \cite{ChangCVPR13}  &  & $\mbox{gPb}_{\mathcal{I}}$ superpixels \cite{ArbelaezMaireFowlkesMalikPAMI11} &  & Our superpixels \tabularnewline
 &  & (117 spx) &  & (101 spx) &  & (101 spx)\tabularnewline
\end{tabular}\hspace*{\fill}
\par\end{centering}
\caption{\label{fig:super-pixels-visualization}Graph based video segmentation
relies on having high quality superpixels/voxels as starting point
(graph nodes). We explore diverse techniques to improve boundary estimates,
which result in better superpixels, which in turn has a significant
impact on final video segmentation.}
\vspace{-3em}
\end{figure*}
\end{abstract}

\section{\label{sec:Introduction}Introduction}

Class-agnostic image and video segmentation have shown to be helpful
in diverse computer vision tasks such as object detection (via object
proposals) \cite{Kraehenbuehl2014Eccv,PontTuset2015ArxivMCG,Humayun2014Cvpr,humayun_ICCV_2015_poise},
semantic video segmentation (as pre-segmentation) \cite{DaiH014},
activity recognition (by computing features on voxels) \cite{TaralovaECCV14},
or scene understanding \cite{Jain2013}.

Both image and video segmentation have seen steady progress recently
leveraging advanced machine learning techniques. A popular and successful
approach consists of modelling segmentation as a graph partitioning
problem \cite{FragkiadakiCVPR12,OchsMalikBroxTPAMI2013,Keuper16Arxiv},
where the nodes represent pixels or superpixels, and the edges encode
the spatio-temporal structure. Previous work focused on solving the
partitioning problem \cite{BroxMalikECCV10,GrundmannKwatraHanEssaCVPR10,PalouCVPR13,Yi2015Iccv},
on the unary and pairwise terms of the graph \cite{Galasso14} and
on the graph construction itself \cite{RenM03,Turaga2009,KhorevaCVPR2015}.

The aim of this paper is to improve video segmentation by focusing
on the graph nodes themselves, the video superpixels. These nodes
are the starting point for unary and pairwise terms, and thus directly
impact the final segmentation quality. Good superpixels for video
segmentation should both be temporally consistent and give high boundary
recall, and, in the case of graph-based video segmentation, for efficient
runtime should enable to use a few superpixels per frame which is
related to high boundary precision. 

Our experiments show that existing classical superpixel/voxel methods
\cite{ChangCVPR13,Achanta2012PamiSlic,VandenberghICCV13} underperform
for graph-based video segmentation and \textcolor{black}{s}uperpixels
built from per-frame boundary estimates are more effective for the
task (see \S\ref{sec:Superpixels-and-supervoxels}). We show that
boundary estimates can be improved when using image cues combined
with object-level cues, and by merging with temporal cues. By fusing
image and time domain cues, we can significantly enhance boundary
estimation in video frames, improve per-frame superpixels, and thus
improve video segmentation. 

In particular we contribute:\vspace{-1em}

\begin{itemize}
\item a comparative evaluation of the importance of the initial superpixels/voxels
for graph-based video segmentations (\S\ref{sec:Superpixels-and-supervoxels}).
\item significantly improved boundary estimates (and thus per-frame superpixels)
by the careful fusion of image (\S\ref{subsec:Image-domain-cues})
and time (\S\ref{subsec:Temporal-cues}) domain cues.
\item the integration of high-level object-related cues into the local image
segmentation processing \textcolor{black}{(\S\ref{subsec:Object-proposals}).}
\item state-of-the-art video segmentation results on the VSB100 \cite{Galasso13}
and BMDS \cite{BroxMalikECCV10} datasets.
\end{itemize}
\vspace{-1.5em}

\section{\label{subsec:Related-work}Related work}

\paragraph{Video segmentation}

Video segmentation can be seen as a clustering problem in the 3D spatial-temporal
volume. Considering superpixels/voxels as nodes, graphs are a natural
way to address video segmentation and there are plenty of approaches
to process the graphs. Most recent and successful techniques include
hybrid generative and discriminative approaches with mixtures of trees
\cite{BadrinarayananIJCV13}, agglomerative methods constructing video
segment hierarchies \cite{GrundmannKwatraHanEssaCVPR10,PalouCVPR13},
techniques based on tracking/propagation of image-initialized solutions
\cite{VandenberghICCV13,ChangCVPR13} and optimization methods based
on Conditional Random Fields \cite{ChengAhujaCVPR12}. We leverage
spectral clustering \cite{ShiMalikTPAMI2000,ng01}, one of the most
successful approaches to video segmentation \cite{FragkiadakiCVPR12,OchsMalikBroxTPAMI2013,ArbelaezMaireFowlkesMalikPAMI11,KhorevaCVPR2015,Keuper16Arxiv}
and consider in our experiments the methods of \cite{Galasso13,Galasso14}.

The above approaches cover various aspects related to graph based
video segmentation. Several papers have addressed the features for
video segmentation \cite{BroxMalikECCV10,GrundmannKwatraHanEssaCVPR10,PalouCVPR13}
and some work has addressed the graph construction \cite{RenM03,Turaga2009}.
While these methods are based on superpixels none of them examines
the quality of the respective superpixels for graph-based video segmentation.
To the best of our knowledge, this work is the first to thoroughly
analyse and advance superpixel methods in the context of video segmentation.

\paragraph{Superpixels/voxels}

We distinguish two groups of superpixel methods. The first one is
the classical superpixel/voxel methods \cite{ChangCVPR13,Achanta2012PamiSlic,VandenberghICCV13,Levinshtein:2009}.
These methods are designed to extract superpixels of homogeneous shape
and size, in order for them to have a regular topology. Having a regular
superpixel topology has shown a good basis for image and video segmentation
\cite{GrundmannKwatraHanEssaCVPR10,PapazoglouFerrariICCV13,BadrinarayananIJCV13,RenM03}. 

The second group are based \textcolor{black}{on boundary estimation
and focus on the image content. They} extract superpixels by building
a hierarchical image segmentation \cite{ArbelaezMaireFowlkesMalikPAMI11,IsolaECCV14,Dollar2015PAMI,PontTuset2015ArxivMCG}
and selecting one level in the hierarchy.\textcolor{black}{{} These
methods generate superpixels of heterogeneous size, that} are typically
fairly accurate on each frame but may jitter over time.\textcolor{black}{{}
Superpixels based on per-frame boundary estimation are employed in
many state-of-the-art video segmentation methods \cite{Galasso14,VazquezreinaECCV10,Jain2013,Yi2015Iccv}.
}\vspace{-1em}

In this work we argue that boundaries based superpixels are more suitable
for graph-based video segmentation, and propose to improve the extracted
superpixels by exploring temporal information such as optical flow
and temporal smoothing.

\paragraph{Image boundaries}

After decades of research on image features and filter banks \cite{ArbelaezMaireFowlkesMalikPAMI11},
most recent methods use machine learning, e.g. decision forests \cite{Dollar2015PAMI,Hallman2015Cvpr},
mutual information \cite{IsolaECCV14}, or convolutional neural networks
\cite{Bertasius2015CvprDeepEdge,Xie_2015_ICCV}. We leverage the latest
trends and further improve them, especially in relation to video data.
\vspace{-0.5em}

\section{\label{sec:Graph-based-video-segmentation}Video segmentation methods}

For our experiments we consider two open source state-of-the-art graph-based
video segmentation methods \cite{Galasso13,Galasso14}. Both of them
rely on superpixels extracted from hierarchical image segmentation
\cite{ArbelaezMaireFowlkesMalikPAMI11}, which we aim to improve.

\paragraph{Spectral graph reduction \cite{Galasso14}}

Our first baseline is composed of three main parts.\\
1.\emph{~Extraction of superpixels.} Superpixels are image-based
pixel groupings which are similar in terms of colour and texture,
extracted by using the state-of-the-art image segmentation of \cite{ArbelaezMaireFowlkesMalikPAMI11}.
These superpixels are accurate but not temporally consistent, as only
extracted per frame. \\
2.\emph{~Feature computation.} Superpixels are compared to their
(spatio-temporal) neighbours and affinities are computed between pairs
of them based on appearance, motion and long term point trajectories
\cite{OchsMalikBroxTPAMI2013}, depending on the type of neighbourhood
(e.g. within a frame, across frames, etc.).\\
3.~\emph{Graph partitioning.} Video segmentation is cast as the grouping
of superpixels into video volumes. \cite{Galasso14} employs either
a spectral clustering or normalised cut formulation for incorporating
a reweighing scheme to improve the performance. 

In our paper we focus on the first part. We show that superpixels
extracted from stronger boundary estimation help to achieve better
segmentation performance without altering the underlying features
or the graph partitioning method.

\paragraph{Segmentation propagation \cite{Galasso13}}

As the second video segmentation method we consider the baseline proposed
in \cite{Galasso13}. This method does greedy matching of superpixels
by propagating them over time via optical flow. This ``simple''
method obtains state-of-the-art performance on VSB100. We therefore
also report how superpixels extracted via hierarchical image segmentation
based on our proposed boundary estimation improve this baseline. \vspace{-2em}

\paragraph{}

\section{\label{sec:Video-segmentation-metrics}Video segmentation evaluation}

\paragraph{VSB100}

We consider for learning and for evaluation the challenging video
segmentation benchmark VSB100 \cite{Galasso13} based on the HD quality
video sequences of \cite{SundbergBroxMaireArbelaezMalikCVPR11}, containing
natural scenes as well as motion pictures, with heterogeneous appearance
and motion. The dataset is arranged into train (40 videos) and test
(60) set. Additionally we split the training set into a training (24)
and validation set (16).

The evaluation in VSB100 is mainly given by:\\
\emph{Precision-recall plots} (BPR, VPR): VSB100 distinguishes a boundary
precision-recall metric (BPR), measuring the per-frame boundary alignment
between a video segmentation solution and the human annotations, and
a volume precision-recall metric (VPR), reflecting the temporal consistency
of the video segmentation result.\\
\emph{Aggregate performance measures} (AP, ODS, OSS): for both BPR
and VPR, VSB100 reports average precision (AP), the area under the
precision-recall curves, and two F-measures where one is measured
at an optimal dataset scale (ODS) and the other at an optimal segmentation
scale (OSS) (where \textquotedbl{}optimal\textquotedbl{} stands for
oracle provided).

\paragraph{BMDS}

To show the generalization of the proposed method we further consider
the Berkeley Motion Segmentation Dataset (BMDS) \cite{BroxMalikECCV10},
which consists of 26 VGA-quality videos, representing mainly humans
and cars. Following prior work \cite{khoreva2014} we use 10 videos
for training and 16 as a test set, and restrict all video sequences
to the first 30 frames. \vspace{-1em}

\section{\label{sec:Superpixels-and-supervoxels}Superpixels and supervoxels}

Graph-based video segmentation methods rely on superpixels to compute
features and affinities. Employing superpixels as pre-processing stage
for video segmentation provides a desirable computational reduction
and a powerful per-frame representation.

Ideally these superpixels have high boundary recall (since one cannot
recover from missing recall), good temporal consistency (to make matching
across time easier), and are as few as possible (in order to reduce
the chances of segmentation errors; to accelerate overall computation
and reduce memory needs). 

In this section we explore which type of superpixels are most suitable
for graph-based video segmentation. 
\begin{figure}[t]
\begin{centering}
\vspace{-1em}
\hspace*{\fill}\subfloat[\label{fig:super-voxels-boundaries-recall}\textcolor{black}{Boundary
recall}]{\begin{centering}
\includegraphics[bb=0bp 0bp 566bp 480bp,width=0.4\textwidth]{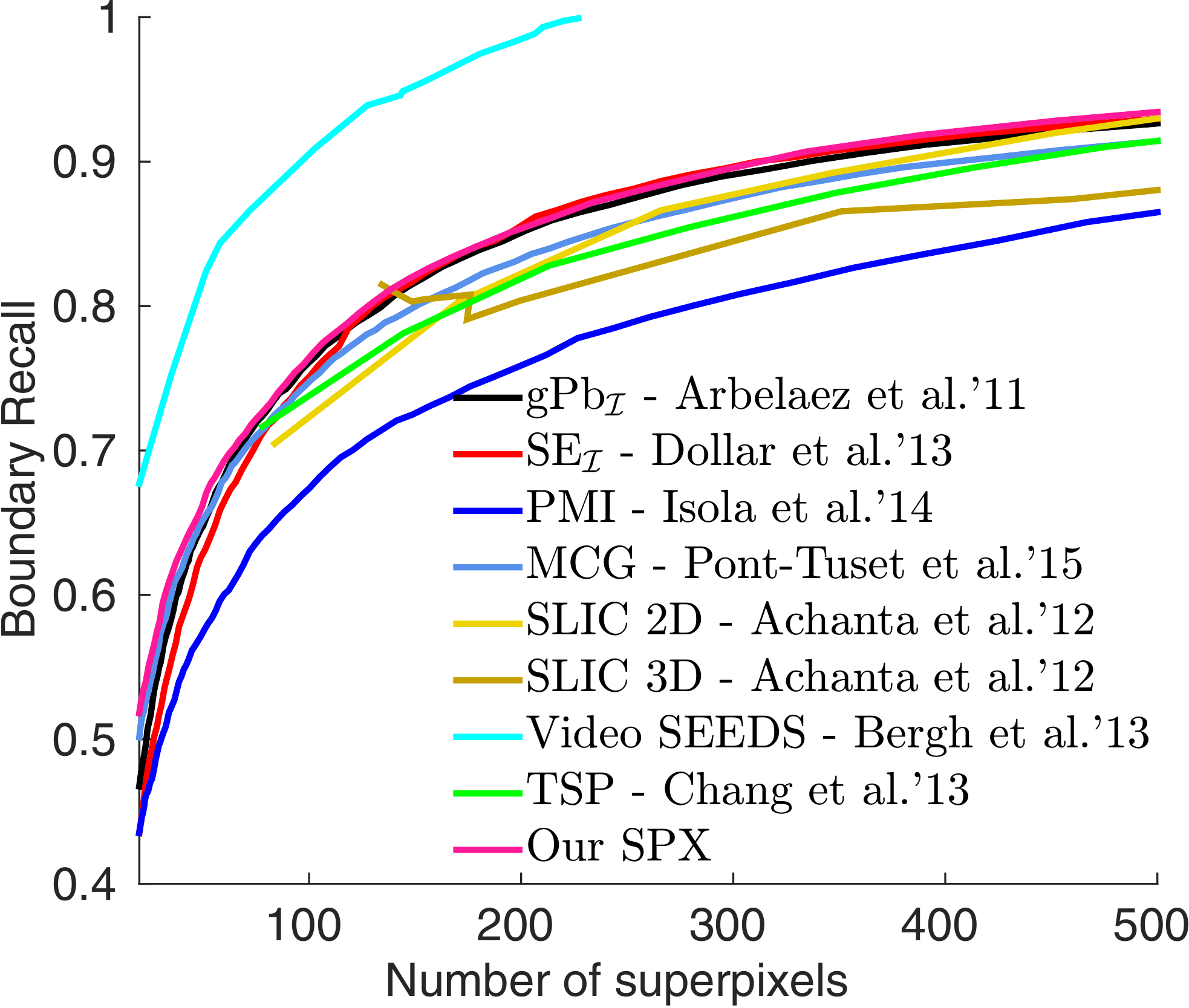}
\par\end{centering}
}\enskip{}\subfloat[\textcolor{red}{\label{fig:super-voxels-boundaries-precision}}\textcolor{black}{Boundary
precision}]{\begin{centering}
\includegraphics[bb=0bp 0bp 576bp 490bp,width=0.4\textwidth]{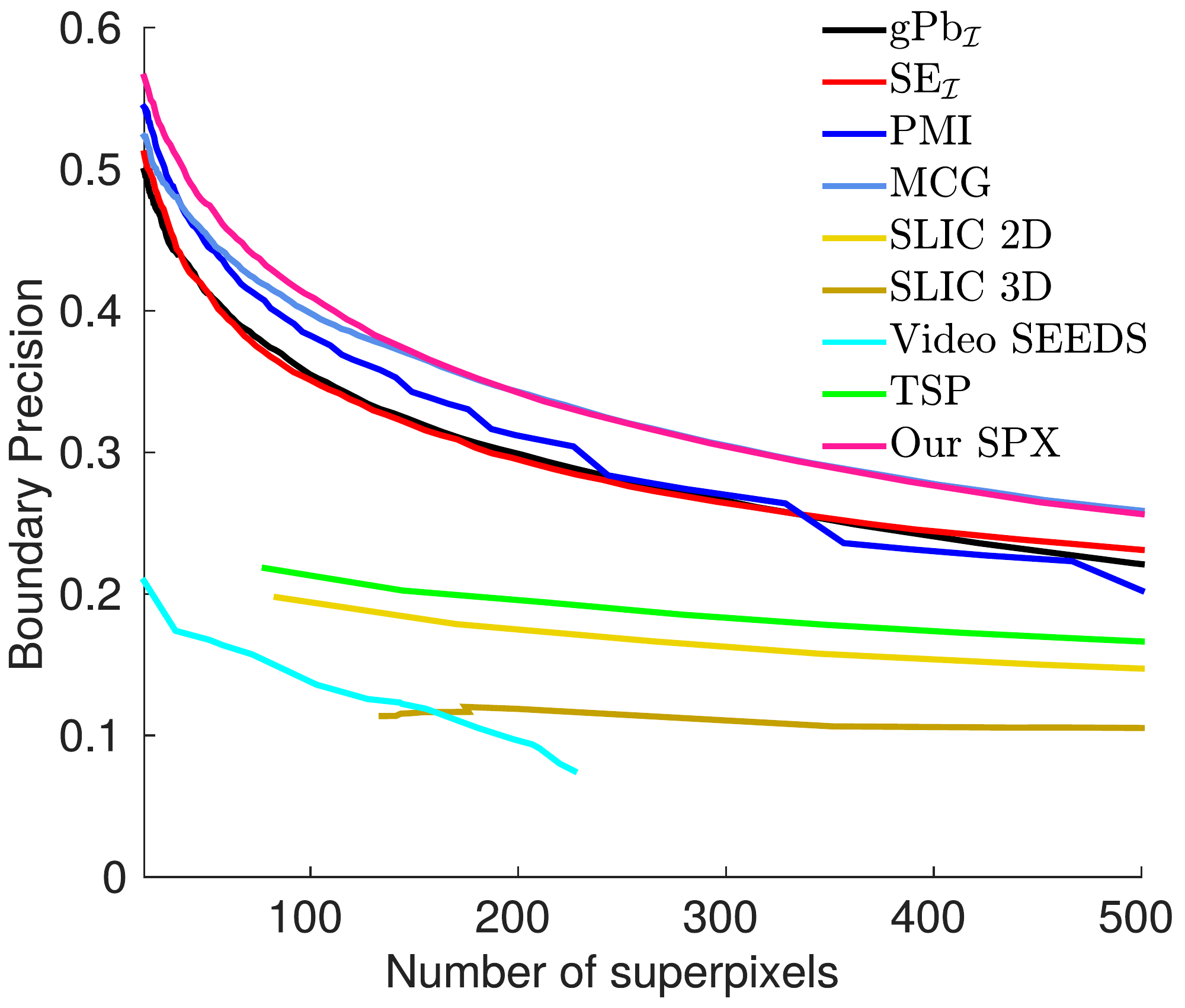}
\par\end{centering}
}\hspace*{\fill}
\par\end{centering}
\begin{centering}
\hspace*{\fill}\subfloat[\label{fig:super-voxels-undersegm-error-1}\textcolor{black}{Under-segmentation
error}]{\begin{centering}
\includegraphics[bb=0bp 0bp 583bp 494bp,width=0.4\textwidth]{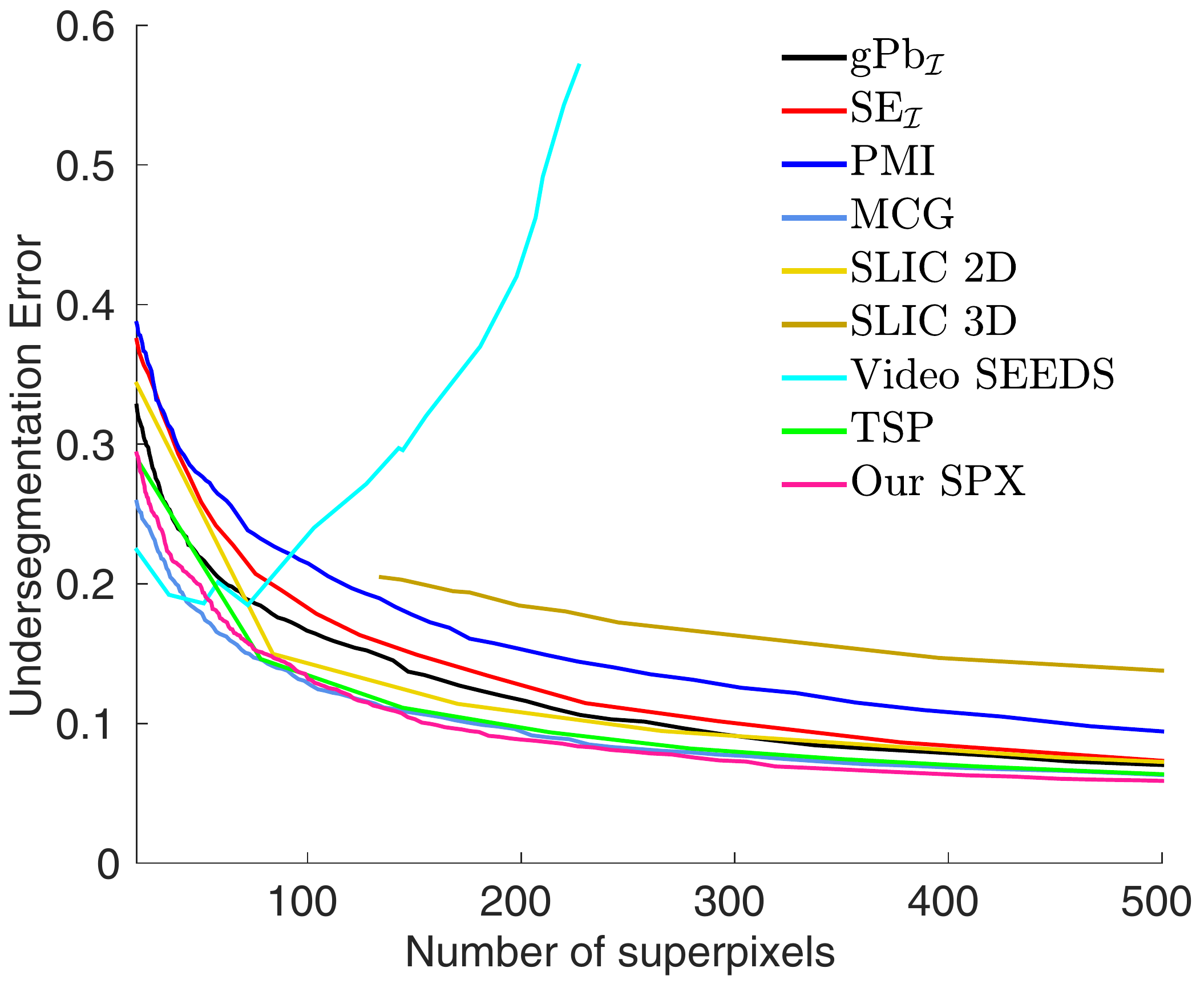}
\par\end{centering}
}\enskip{}\subfloat[\label{fig:super-voxels-for-video-segmentation}\textcolor{black}{BPR
of video segmentation}]{\begin{centering}
\includegraphics[bb=0bp 0bp 576bp 490bp,width=0.4\textwidth]{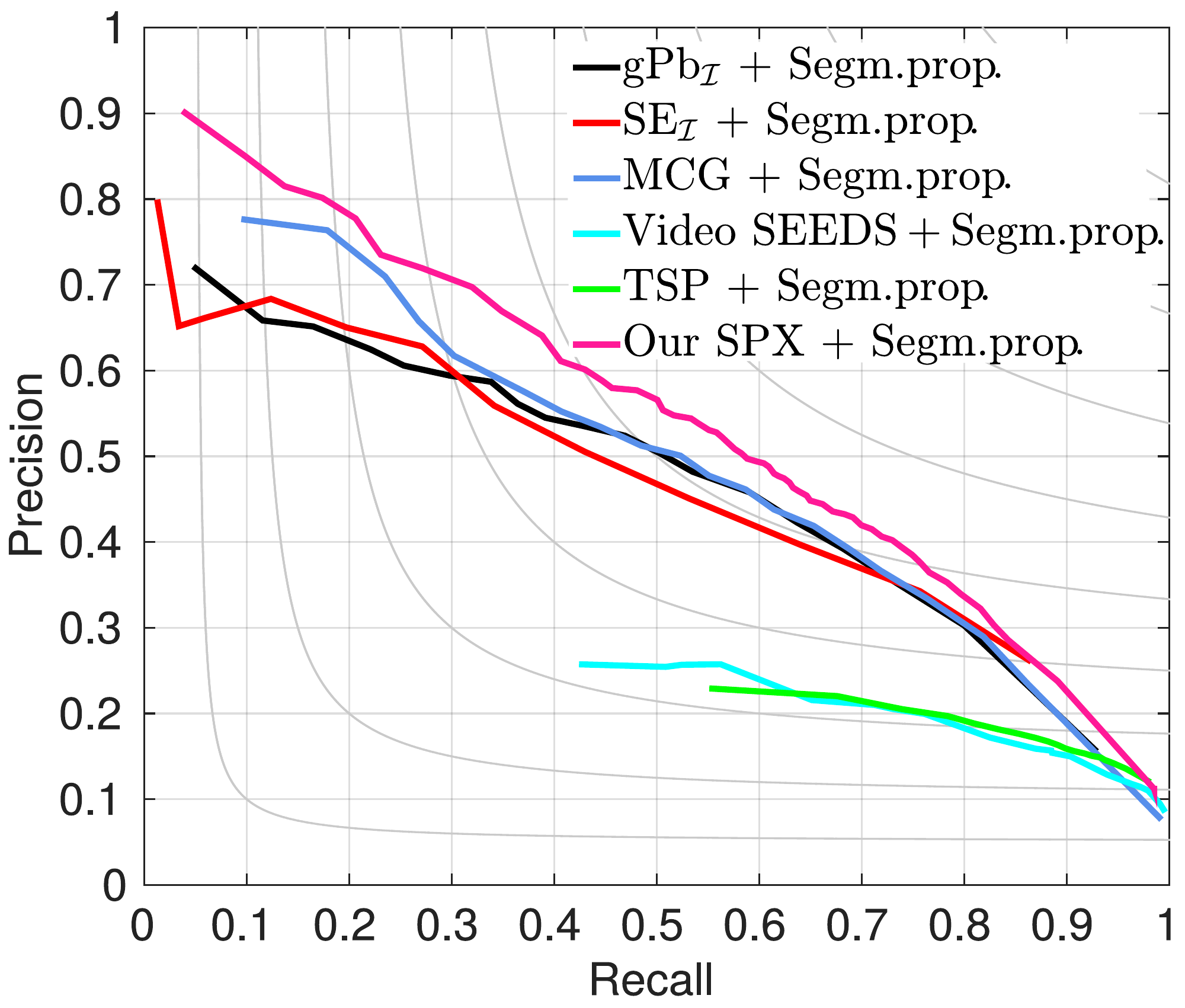}
\par\end{centering}
}\hspace*{\fill}
\par\end{centering}
\caption{\label{fig:comparing-supervoxels}Comparison of different superpixel/voxel
methods, and their use for video segmentation. VSB100 validation set.
SPX: superpixels. \textcolor{black}{Segm. prop.: segmentation propagation
\cite{Galasso13} (see \S\ref{sec:Graph-based-video-segmentation}).}}
\vspace{0em}
\end{figure}

\paragraph{\textcolor{black}{Superpixel/voxel methods}}

\textcolor{black}{Many superpixel/voxel methods have been explored
in the past. We consider the most promising ones in the experiments
of Figure \ref{fig:comparing-supervoxels}. SLIC 2D/3D \cite{Achanta2012PamiSlic}
is a classic method to obtain superpixels via iterative clustering
(in space and space-time domain). TSP \cite{ChangCVPR13} extends
SLIC to explicitly model temporal dynamics. Video SEEDS \cite{VandenberghICCV13}
is similar to SLIC 3D, but uses an alternative optimization strategy.
Other than classic superpixel/voxel methods we also consider superpixels
generated from per-frame hierarchical segmentation based on boundary
detection (}ultrametric contour map\textcolor{black}{s \cite{ArbelaezMaireFowlkesMalikPAMI11}).
We include $\mbox{gPb}_{\mathcal{I}}$ \cite{ArbelaezMaireFowlkesMalikPAMI11},
$\mbox{SE}_{\mathcal{I}}$ \cite{Dollar2015PAMI}, PMI \cite{IsolaECCV14}
and MCG \cite{PontTuset2015ArxivMCG} as sources of boundary estimates. }

\paragraph{\textcolor{black}{Superpixel evaluation}}

\textcolor{black}{We compare superpixels by evaluating the recall
and precision of boundaries and the under-segmentation error \cite{NeubertBMVC13}
as functions of the average number of superpixels per frame. We also
use some of them directly for video segmentation (Figure \ref{fig:super-voxels-for-video-segmentation}).
We evaluate (use) all methods on a frame by frame basis; supervoxel
methods are expected to provide more temporally consistent segmentations
than superpixel methods.}

\paragraph{\textcolor{black}{Results}}

\textcolor{black}{Boundary recall (Figure \ref{fig:super-voxels-boundaries-recall})
is comparable for most methods. Video SEEDS is an outlier, showing
very high recall, but low boundary precision (\ref{fig:super-voxels-boundaries-precision})
and high under-segmentation error (\ref{fig:super-voxels-undersegm-error-1}).
$\mbox{gPb}_{\mathcal{I}}$ and $\mbox{SE}_{\mathcal{I}}$ reach the
highest boundary recall with fewer superpixels. Per-frame boundaries
based superpixels perform better than classical superpixel methods
on boundary precision (\ref{fig:super-voxels-boundaries-precision}).
From these figures one can see the conflicting goals of having high
boundary recall, high precision, and few superpixels.}\\
\textcolor{black}{We additionally evaluate the superpixel methods
using a region-based metric: under-segmentation error \cite{NeubertBMVC13}.
Similar to the boundary results, the curves are clustered in two groups:
TSP-like and $\mbox{gPb}_{\mathcal{I}}$-like quality methods, where
the latter underperform due to the heterogeneous shape and size of
superpixels (\ref{fig:super-voxels-undersegm-error-1}). }\vspace{-1em}

\textcolor{black}{Figure \ref{fig:super-voxels-for-video-segmentation}
shows the impact of superpixels for video segmentation using the baseline
method \cite{Galasso13}. We pick TSP as a representative superpixel
method (fair quality on all metrics), Video SEEDS as an interesting
case (good boundary recall, bad precision), $\mbox{SE}_{\mathcal{I}}$
and MCG as good boundary estimation methods, and the baseline $\mbox{gPb}_{\mathcal{I}}$
(used in \cite{Galasso13}). Albeit classical superpixel methods have
lower under-segmentation error than boundaries based superpixels,
when applied for video segmentation the former underperform (both
on boundary and volume metrics), as seen in Figure \ref{fig:super-voxels-for-video-segmentation}.
Boundary quality measures seem to be a good proxy to predict the quality
of superpixels for video segmentation. Both in boundary precision
and recall metrics having stronger initial superpixels leads to better
results. }

\paragraph{\textcolor{black}{Intuition}}

\textcolor{black}{Figure \ref{fig:super-pixels-visualization} shows
a visual comparison of TSP superpixels versus $\mbox{gPb}_{\mathcal{I}}$
superpixels (both generated with a similar number of superpixels).
By design, most classical superpixel methods have a tendency to generate
superpixels of comparable size. When requested to generate fewer superpixels,
they need to trade-off quality versus regular size. Methods based
on hierarchical segmentation (such as $\mbox{gPb}_{\mathcal{I}}$)
generate superpixels of heterogeneous sizes and more likely to form
semantic regions. For a comparable number of superpixels techniques
based on image segmentation have more freedom to provide better superpixels
for graph-based video segmentation than classical superpixel methods.}

\paragraph{\textcolor{black}{Conclusion}}

\textcolor{black}{Based both on quality metrics and on their direct
usage for graph-based video segmentation, boundary based superpixels
extracted via hierarchical segmentation are more effective than the
classical superpixel methods in the context of video segmentation.
The hierarchical segmentation is fully defined by the estimated boundary
probability, thus better boundaries lead to better superpixels, which
in turn has a significant impact on final video segmentation. In the
next sections we discuss how to improve boundary estimation for video.}

\section{\label{sec:Improving-image-boundaries}Improving image boundaries}

To improve the boundary based superpixels fed into video segmentation
we seek to make best use of the information available on the videos.
We first improve boundary estimates using each image frame separately
(\S\ref{subsec:Image-domain-cues}) and then consider the temporal
dimension (\S\ref{subsec:Temporal-cues}).

\subsection{\label{subsec:Image-domain-cues}Image domain cues}

A classic boundary estimation method (often used in video segmentation)
is $\mbox{gPb}_{\mathcal{I}}$ \cite{ArbelaezMaireFowlkesMalikPAMI11}
($\mathcal{I}$: image domain), we use it as a reference point for
boundary quality metrics. In our approach we propose to use $\mbox{SE}_{\mathcal{I}}$
(``structured edges'') \cite{Dollar2015PAMI}. We also considered
the convnet based boundary detector \cite{Xie_2015_ICCV}. \textcolor{black}{However
employing} boundaries of \cite{Xie_2015_ICCV} to close the contours
and construct \textcolor{black}{per-frame hierarchical segmentation
results in the performance similar to }$\mbox{SE}_{\mathcal{I}}$
and significantly longer training time. Therefore in our system we
employ $\mbox{SE}_{\mathcal{I}}$ due to its speed and good quality.

\vspace{-1em}

\subsubsection{\label{subsec:Object-proposals}Object proposals}

Methods such as $\mbox{gPb}_{\mathcal{I}}$ and $\mbox{SE}_{\mathcal{I}}$
use bottom-up information even though boundaries annotated by humans
in benchmarks such as BSDS500 or VSB100 often follow object boundaries.
In other words, an oracle having access to ground truth semantic object
boundaries should allow to improve boundary estimation (in particular
on the low recall region of the BPR curves). Based on this intuition
we consider using segment-level object proposal (OP) methods to improve
initial boundary estimates ($\mbox{SE}_{\mathcal{I}}$). Object proposal
methods \cite{Kraehenbuehl2014Eccv,PontTuset2015ArxivMCG,Humayun2014Cvpr,humayun_ICCV_2015_poise}
aim at generating a set of candidate segments likely to have high
overlap with true objects. Typically such methods reach $\sim\negmedspace80\%$
object recall with $10^{3}$ proposals per image.

Based on initial experiments we found that the following simple approach
obtains good boundary estimation results in practice. Given a set
of object proposal segments generated from an initial boundary estimate,
we average the contours of each segment. Pixels that are boundaries
to many object proposals will have high probability of boundary; pixels
rarely members of a proposal boundary will have low probability. With
this approach, the better the object proposals, the closer we are
to the mentioned oracle case.

We evaluated multiple proposals methods \cite{Kraehenbuehl2014Eccv,PontTuset2015ArxivMCG,Humayun2014Cvpr}
and found RIGOR \cite{Humayun2014Cvpr} to be most effective for this
use (\S\ref{subsec:Results-appearance-only}). To the best of our
knowledge this is the first time an object proposal method is used
to improve boundary estimation. We name the resulting boundary map
$\mbox{OP}\left(\mbox{SE}_{\mathcal{I}}\right)$.\vspace{-1em}

\subsubsection{Globalized probability of boundary}

A key ingredient of the classic $\mbox{gPb}_{\mathcal{I}}$ \cite{ArbelaezMaireFowlkesMalikPAMI11}
method consists on ``globalizing boundaries''. The most salient
boundaries are highlighted by computing a weighted sum of the spatial
derivatives of the first few eigenvectors of an affinity matrix built
based on an input probability of boundary. The affinity matrix can
be built either at the pixel or superpixel level. The resulting boundaries
are named ``spectral'' \textcolor{black}{probability of boundary,
$\mbox{sPb}\left(\cdot\right)$. We employ the fast implementation
from \cite{PontTuset2015ArxivMCG}.}

\textcolor{black}{Albeit well known, such a globalization step is
not considered by the latest work on boundary estimation (e.g. \cite{Dollar2015PAMI,Bertasius2015CvprDeepEdge}).
Since we compute boundaries at a single-scale, $\mbox{sPb}\left(\mbox{\ensuremath{\mbox{SE}_{\mathcal{I}}}}\right)$
is comparable to the SCG results in \cite{PontTuset2015ArxivMCG}.}

\subsubsection{\label{subsec:Re-training}Re-training}

Methods such as $\mbox{SE}_{\mathcal{I}}$ are trained and tuned for
the BSDS500 image segmentation dataset \cite{ArbelaezMaireFowlkesMalikPAMI11}.
Given that VSB100 \cite{Galasso13} is larger and arguably more relevant
to the video segmentation task than BSDS500, we retrain $\mbox{SE}_{\mathcal{I}}$
(and RIGOR) for this task. In the following sections we report results
of our system trained over BSDS500, or with VSB100. We will also consider
using input data other than an RGB image (\S \ref{subsec:Optical-flow}).

\vspace{-1em}

\subsubsection{\label{subsec:Merging-cues}Merging cues}

After obtaining complementary probabilities of boundary maps (e.g.
$\mbox{OP}\left(\mbox{SE}_{\mathcal{I}}\right)$, $\mbox{sPb}\left(\mbox{\ensuremath{\mbox{SE}_{\mathcal{I}}}}\right)$,
etc.), we want to combine them effectively. Naive averaging is inadequate
because boundaries estimated by different methods do not have pixel-perfect
alignment amongst each other. Pixel-wise averaging or maxing leads
to undesirable double edges (negatively affecting boundary precision). 

To solve this issue we use the grouping technique from \cite{PontTuset2015ArxivMCG}
which proposes to first convert the boundary estimate into a hierarchical
segmentation, and then to align the segments from different methods.
Note that we do not use the multi-scale part of \cite{PontTuset2015ArxivMCG}.
Unless otherwise specified all cues are averaged with equal weight.
We use the sign ``$+$'' to indicate such merges.

\vspace{-1em}

\subsubsection{\label{subsec:Results-appearance-only}Boundary results when using
image domain cues}

\begin{figure}[t]
\centering{}%
\begin{minipage}[t]{0.48\textwidth}%
\begin{center}
\includegraphics[bb=0bp 0bp 598bp 490bp,width=1\columnwidth]{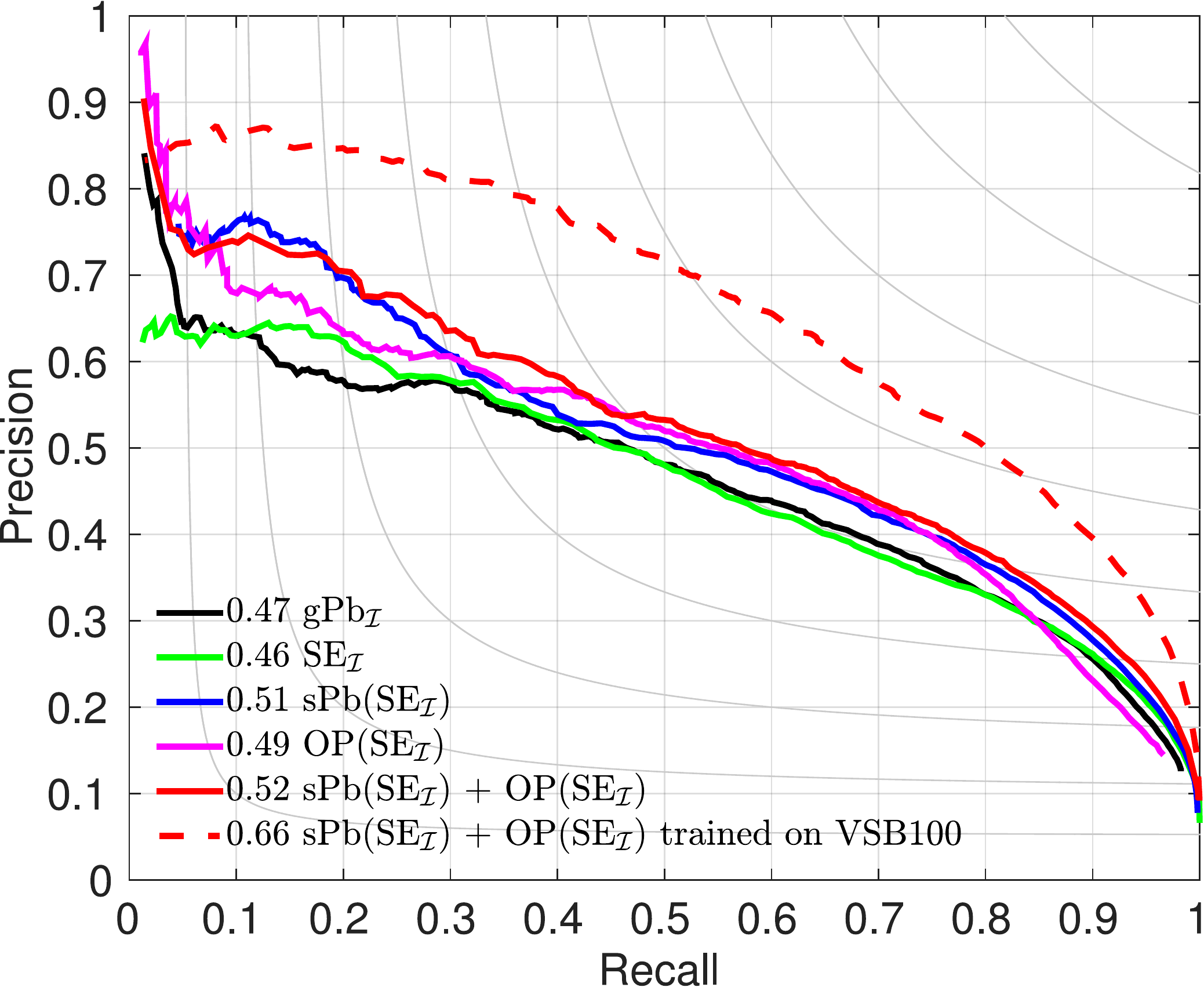}\caption{\label{fig:appeareance-boundaries-validation-set} Progress when integrating
various image domain cues (\S\ref{subsec:Image-domain-cues}) in
terms of BPR on VSB100 validation set.}
\par\end{center}%
\end{minipage}\hspace*{\fill}%
\begin{minipage}[t]{0.48\textwidth}%
\begin{center}
\includegraphics[bb=0bp 0bp 601bp 490bp,width=1\columnwidth]{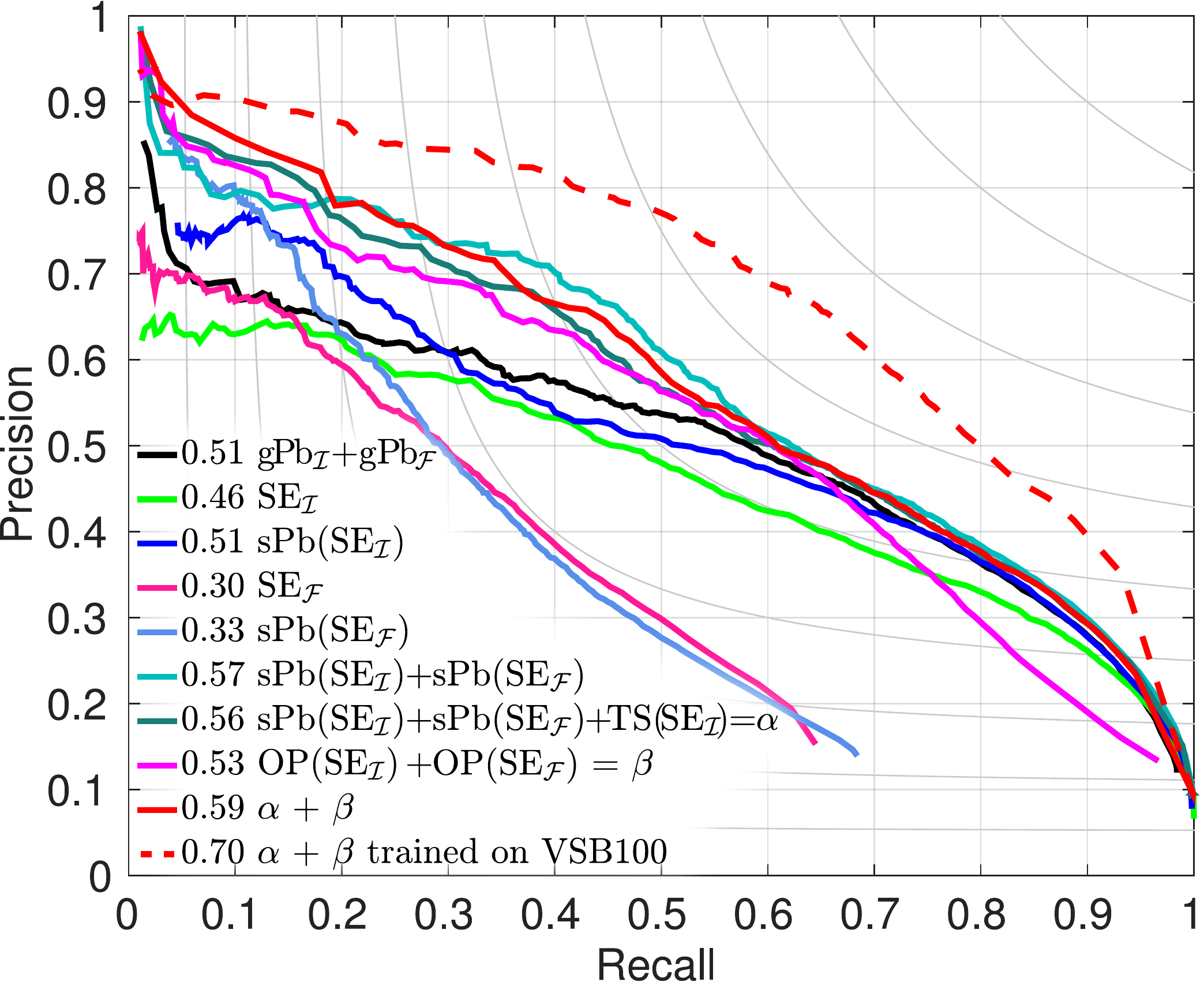}\caption{\label{fig:flow-boundaries-validation-set}Progress when integrating
image and time domain cues (\S\ref{subsec:Temporal-cues}) in terms
of BPR on VSB100 validation set.}
\par\end{center}%
\end{minipage}\vspace{-0.5em}
\end{figure}
Figure \ref{fig:appeareance-boundaries-validation-set} reports results
when using the different image domain cues, evaluated over the VSB100
validation set. The $\mbox{gPb}_{\mathcal{I}}$ baseline obtains 47\%
AP, while $\mbox{SE}_{\mathcal{I}}$ (trained on BSDS500) obtains
46\%. Interestingly, boundaries based on object proposals $\mbox{OP}\left(\mbox{\ensuremath{\mbox{SE}_{\mathcal{I}}}}\right)$
from RIGOR obtain a competitive 49\%, and, as expected, provide most
gain in the high precision region of BPR. Globalization $\mbox{sPb}\left(\mbox{\ensuremath{\mbox{SE}_{\mathcal{I}}}}\right)$
improves results to 51\% providing a homogeneous gain across the full
recall range. Combining $\mbox{sPb}\left(\mbox{\ensuremath{\mbox{SE}_{\mathcal{I}}}}\right)$
and $\mbox{OP}\left(\mbox{\ensuremath{\mbox{SE}_{\mathcal{I}}}}\right)$
obtains 52\%. After retraining $\mbox{SE}_{\mathcal{I}}$ on VSB100
we obtain our best result of 66\% AP (note that all cues are affected
by re-training $\mbox{SE}_{\mathcal{I}}$).

\paragraph{Conclusion}

Even when using only image domain cues, large gains can be obtained
over the standard $\mbox{gPb}_{\mathcal{I}}$ baseline.

\subsection{\label{subsec:Temporal-cues}Temporal cues}

The results of \S\ref{subsec:Image-domain-cues} ignore the fact
that we are processing a video sequence. In the next sections we describe
two different strategies to exploit the temporal dimension.

\vspace{-1em}

\subsubsection{\label{subsec:Optical-flow}Optical flow}

We propose to improve boundaries for video by employing optical flow
cues. We use the state-of-the-art EpicFlow \cite{Revaud2015Cvpr}
algorithm, which we feed with our $\mbox{\ensuremath{\mbox{SE}_{\mathcal{I}}}}$
boundary estimates. 

Since optical flow is expected to be smooth across time, if boundaries
are influenced by flow, they will become more temporally consistent.
Our strategy consists of computing boundaries directly over the forward
and backward flow map, by applying SE over the optical flow magnitude
(similar to one of the cues used in \cite{Fragkiadaki2014Arxiv}).
We name the resulting boundaries map $\mbox{\mbox{SE}}_{\mathcal{F}}$
($\mathcal{F}$: optical flow). Although the flow magnitude disregards
the orientation information from the flow map, in practice discontinuities
in magnitude are related to changes in flow direction.

We then treat $\mbox{\mbox{SE}}_{\mathcal{F}}$ similarly to $\mbox{\ensuremath{\mbox{SE}_{\mathcal{I}}}}$
and compute $\mbox{OP}\left(\mbox{\ensuremath{\mbox{SE}_{\mathcal{F}}}}\right)$
and $\mbox{sPb}\left(\mbox{\ensuremath{\mbox{SE}_{\mathcal{F}}}}\right)$
over it. All these cues are finally merged using the method described
in \S\ref{subsec:Merging-cues}.

\vspace{-1em}

\subsubsection{\label{subsec:Time-smoothing}Time smoothing}

\textcolor{black}{The goal of our new boundaries based superpixels}\textcolor{red}{{}
}\textcolor{black}{is not only high recall, but also good temporal
consistency across frames. A naive way to improve temporal smoothness
of boundaries consists of averaging boundary maps of different frames
over a sliding window; differences across frames would be smoothed
out, but at the same time double edge artefacts (due to motion) would
appear (reduced precision).}

\textcolor{black}{We propose to improve temporal consistency by doing
a sliding window average across boundary maps of several adjacent
frames. For each frame $t$, instead of naively transferring boundary
estimates from one frame to the next, we warp frames $t_{\pm i}$
using optical flow with respect to frame $t$; thus reducing double
edge artefacts. For each frame $t$ we treat warped boundaries from
frames $t_{\pm i}$ as additional cues, and merge them using the same
mechanism as in }\S\textcolor{black}{\ref{subsec:Merging-cues}.
This merging mechanism is suitable to further reduce the double edges
issue.}

\subsubsection{\label{subsec:Boundary-result-with-time}Boundary results when using
temporal cues}

The curves of Figure \ref{fig:flow-boundaries-validation-set} show
the improvement gained from optical flow and temporal smoothing.

\paragraph{Optical flow}

Figure \ref{fig:flow-boundaries-validation-set} shows that on its
own flow boundaries are rather weak ($\mbox{\ensuremath{\mbox{SE}_{\mathcal{F}}}}$,
$\mbox{sPb}\left(\mbox{\ensuremath{\mbox{SE}_{\mathcal{F}}}}\right)$),
but they are quite complementary to image domain cues ($\mbox{sPb}\left(\mbox{\ensuremath{\mbox{SE}_{\mathcal{I}}}}\right)$
versus $\mbox{sPb}\left(\mbox{\ensuremath{\mbox{SE}_{\mathcal{I}}}}\right)\negmedspace+\negmedspace\mbox{sPb}\left(\mbox{\ensuremath{\mbox{SE}_{\mathcal{F}}}}\right)$).

\paragraph{Temporal smoothing}

Using temporal smoothing ($\mbox{ sPb}\left(\mbox{\ensuremath{\mbox{SE}_{\mathcal{I}}}}\right)\negmedspace+\negmedspace\mbox{sPb}\left(\mbox{\ensuremath{\mbox{SE}_{\mathcal{F}}}}\right)\negthinspace+\negmedspace\mbox{TS}\left(\mbox{\ensuremath{\mbox{SE}_{\mathcal{I}}}}\right)\negmedspace=\negmedspace\alpha$)
leads to a minor drop in boundary precision, in comparison with $\mbox{sPb}\left(\mbox{\ensuremath{\mbox{SE}_{\mathcal{I}}}}\right)\negmedspace+\negmedspace\mbox{sPb}\left(\mbox{\ensuremath{\mbox{SE}_{\mathcal{F}}}}\right)$
in Figure \ref{fig:flow-boundaries-validation-set}. \textcolor{black}{It
should be noted that there is an inherent tension between improving
temporal smoothness of the boundaries and having better accuracy on
a frame by frame basis. Thus we aim for the smallest negative impact
on BPR. }In our preliminary experiments the key for temporal smoothing
was to use the right merging strategy (\S\ref{subsec:Merging-cues}).
We expect temporal smoothing to improve temporal consistency.

\paragraph{Object proposals}

Adding $\mbox{OP}\left(\mbox{\ensuremath{\mbox{SE}_{\mathcal{F}}}}\right)$
over $\mbox{OP}\left(\mbox{\ensuremath{\mbox{SE}_{\mathcal{I}}}}\right)$
also improves BPR (see $\mbox{OP}\left(\mbox{\ensuremath{\mbox{SE}_{\mathcal{F}}}}\right)+$$\mbox{OP}\left(\mbox{\ensuremath{\mbox{SE}_{\mathcal{I}}}}\right)\negmedspace=$$\beta$
in Figure \ref{fig:flow-boundaries-validation-set}), particularly
in the high-precision area. Merging it with other cues helps to push
BPR for our final frame-by-frame result.

\paragraph{Combination and re-training}

Combining all cues together improves the BPR metric with respect to
only using appearance cues, we reach $59\%\ \mbox{AP}$ versus $52\%$
with appearance only (see \S\ref{subsec:Results-appearance-only}).
This results are better than the $\mbox{gPb}_{\mathcal{I}}\negmedspace+\negmedspace\mbox{gPb}_{\mathcal{F}}$
baseline ($51\%\ \mbox{AP}$, used in \cite{Galasso14}). \\
Similar to the appearance-only case, re-training over VSB100 gives
an important boost ($70\%\ \mbox{AP}$). In this case not only $\mbox{SE}_{\mathcal{I}}$
is re-trained but also $\mbox{SE}_{\mathcal{F}}$ (over EpicFlow).

\begin{figure*}[t]
\begin{centering}
\vspace{-2em}
\hspace*{\fill}\subfloat[\label{fig:bpr-validation-set}BPR on validation set]{\begin{centering}
\includegraphics[bb=0bp 0bp 598bp 490bp,width=0.35\columnwidth]{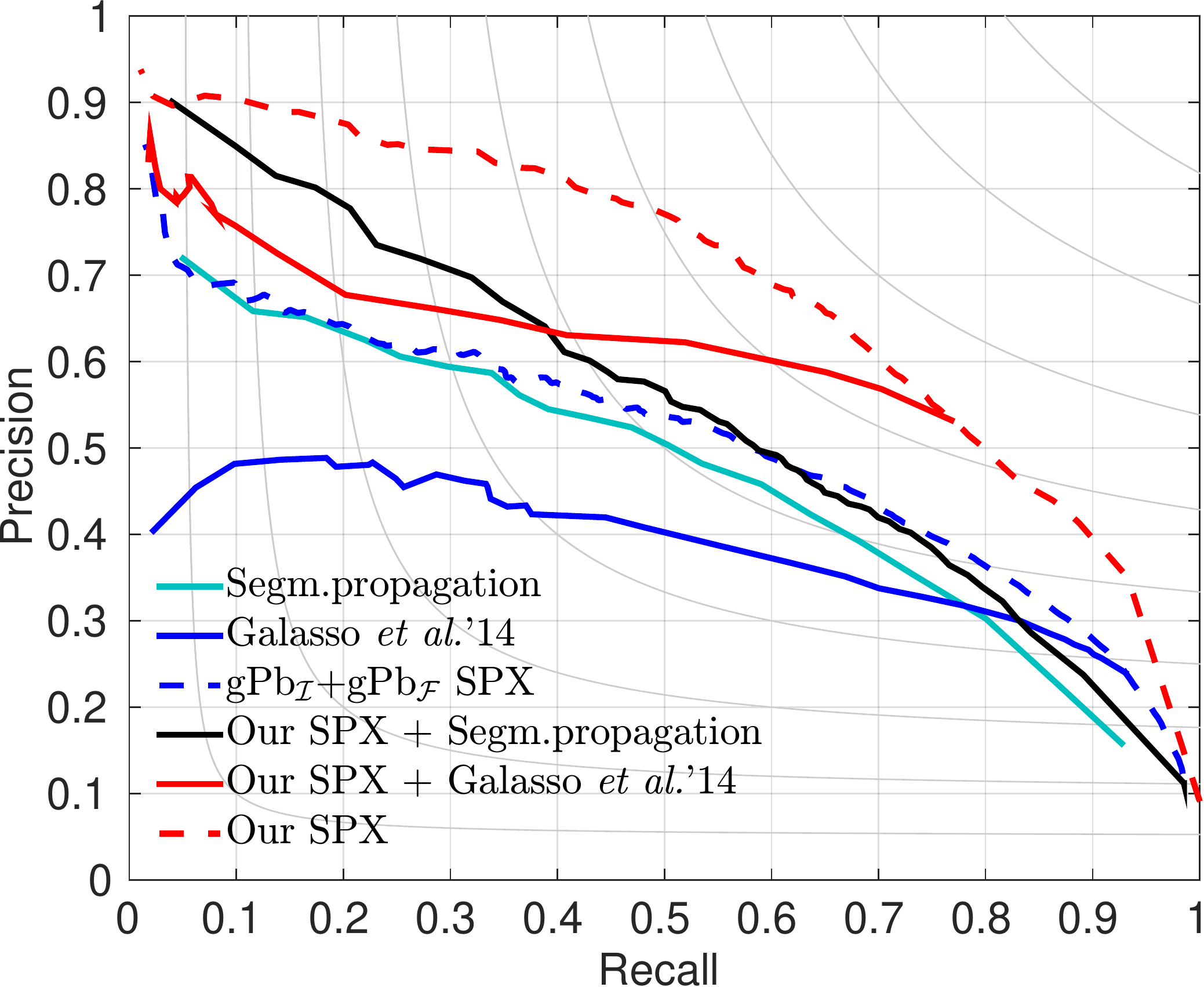}
\par\end{centering}
}\subfloat[\label{fig:vpr-validation-set}VPR on validation set]{\begin{centering}
\includegraphics[bb=0bp 0bp 599bp 490bp,width=0.35\columnwidth]{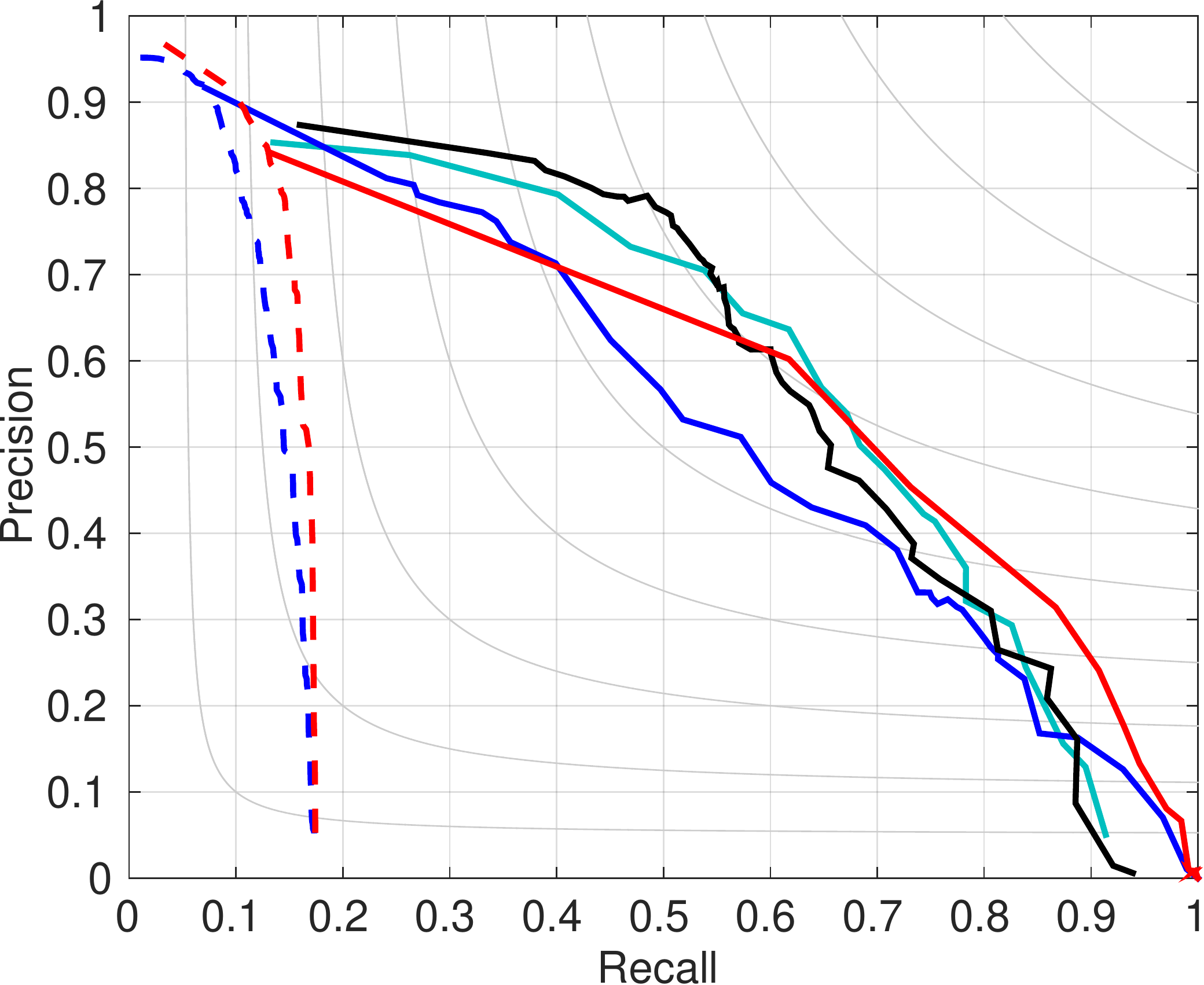}
\par\end{centering}
}\subfloat[\label{fig:per-video-validation}OSS per video]{\begin{centering}
\begin{minipage}[b]{0.24\columnwidth}%
\includegraphics[bb=0bp 0bp 591bp 364bp,width=1\textwidth]{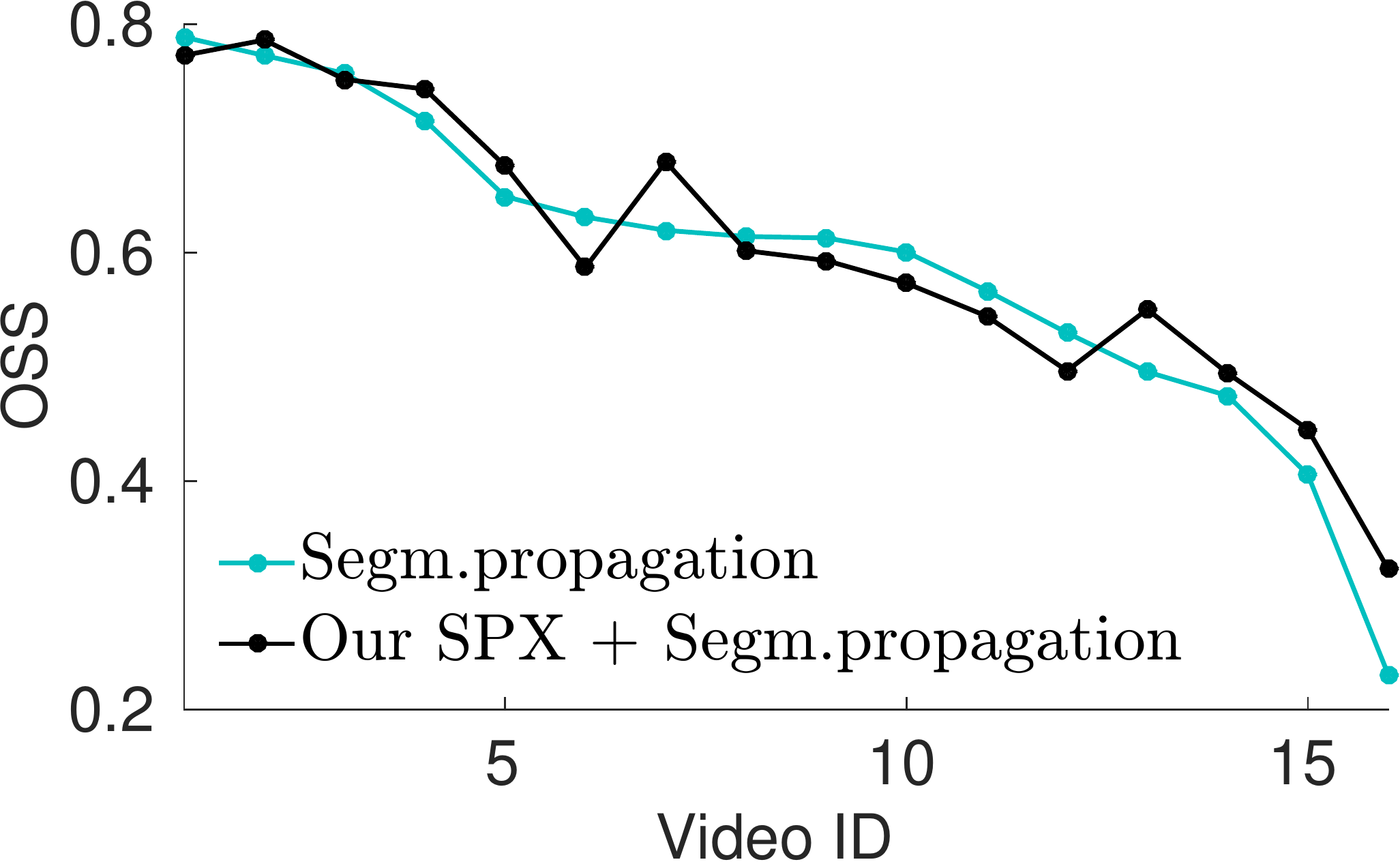}

\includegraphics[bb=0bp 0bp 591bp 364bp,width=1\textwidth]{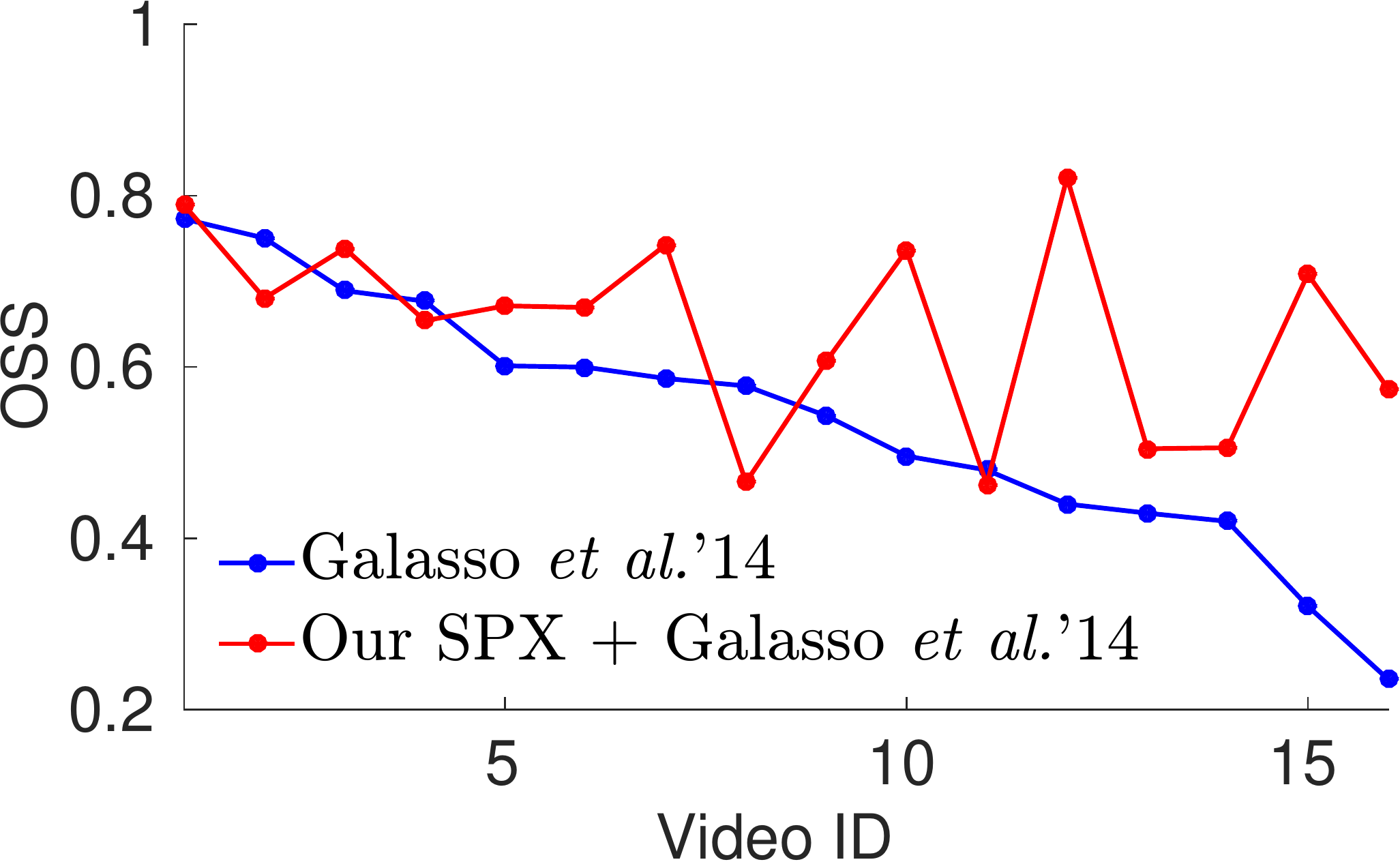}%
\end{minipage}
\par\end{centering}
}\hspace*{\fill}\vspace{-0.5em}
\par\end{centering}
\caption{\label{fig:video-segmentation-validation-set}\textcolor{black}{VSB100}
validation set results of different video segmentation methods. Dashed
lines indicate only frame-by-frame processing (see \S\ref{subsec:Validation-set-video-segmentation-results}
for details).}
\end{figure*}

\textcolor{black}{Figure \ref{fig:comparing-supervoxels} compares
superpixels extracted from the proposed method ($\alpha\negmedspace+\negmedspace\beta$
model without re-training for fair comparison) with other methods.
Our method reaches top results on both boundary precision and recall.
Unless otherwise specified, all following ``Our SPX'' results correspond
to superpixels generated from the hierarchical image segmentation
\cite{ArbelaezMaireFowlkesMalikPAMI11} based on the proposed boundary
estimation $\alpha\negmedspace+\negmedspace\beta$ re-trained on VSB100.}

\paragraph{Conclusion}

Temporal cues are effective at improving the boundary detection for
video sequences. Because we use multiple ingredients based on machine
learning, training on VSB100 significantly improves quality of boundary
estimates on a per-frame basis (BPR). 

\section{\label{sec:Video-segmentation-results}Video segmentation results}

In this section we show results for the state-of-the-art video segmentation
methods \cite{Galasso13,Galasso14} with superpixels \textcolor{black}{extracted
from the proposed} boundary estimation. So far we have only evaluated
boundaries of frame-by-frame hierarchical segmentation. For all further
experiments we will use the best performing model trained on VSB100,
which uses image domain and temporal cues, proposed in \S\ref{sec:Improving-image-boundaries}
(we refer to ($\alpha+\beta$) model, see Figure \ref{fig:flow-boundaries-validation-set}).
Superpixels extracted from our boundaries help to improve video segmentation
and generalizes across different datasets. 
\begin{figure*}[t]
\begin{centering}
\vspace{-1em}
\par\end{centering}
\begin{centering}
\hspace*{\fill}\subfloat[\label{fig:bpr-test-set}BPR on test set]{\begin{centering}
\includegraphics[width=0.47\columnwidth]{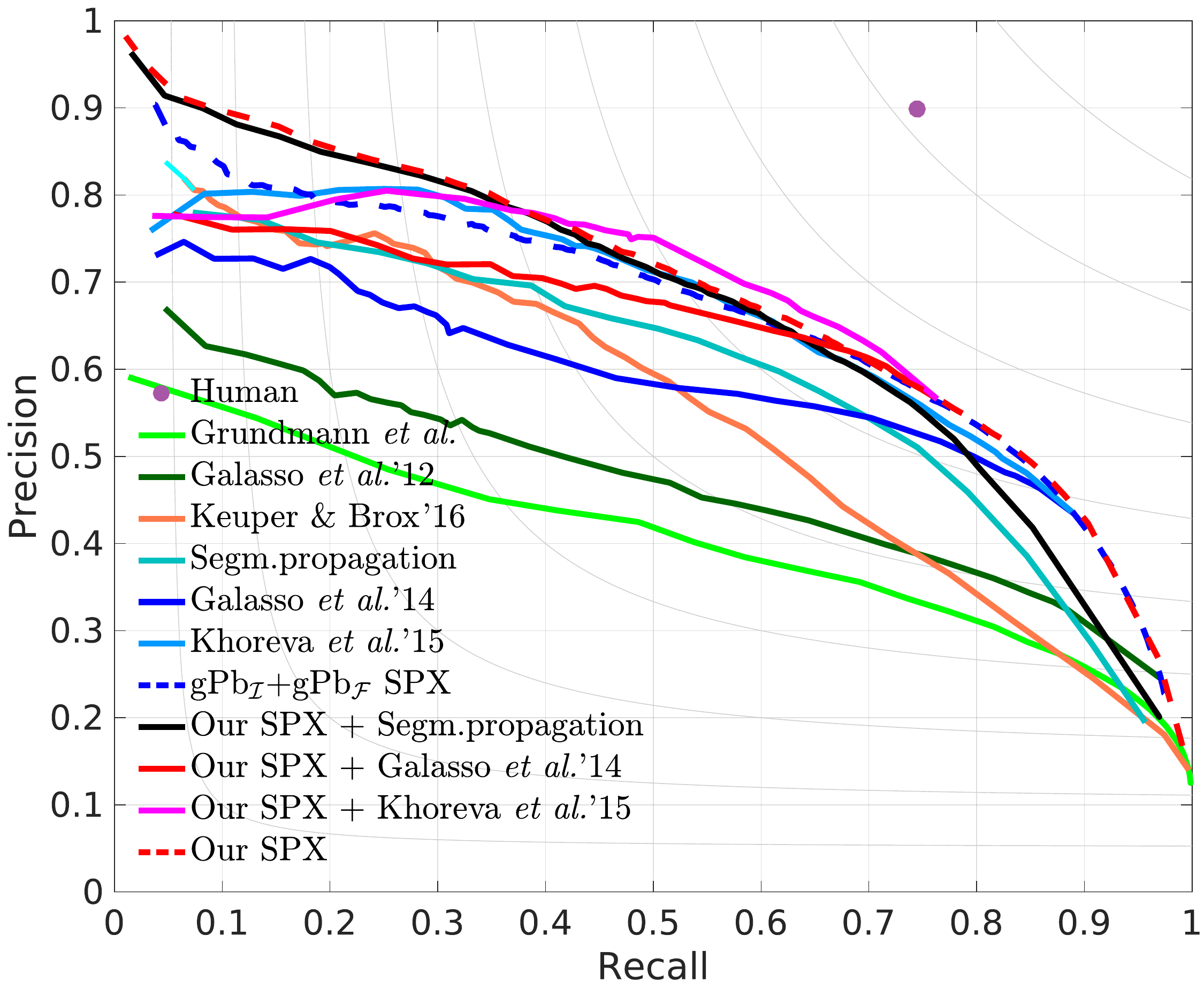}
\par\end{centering}
}\subfloat[\label{fig:vpr-test-set}VPR on test set]{\begin{centering}
\includegraphics[width=0.47\columnwidth]{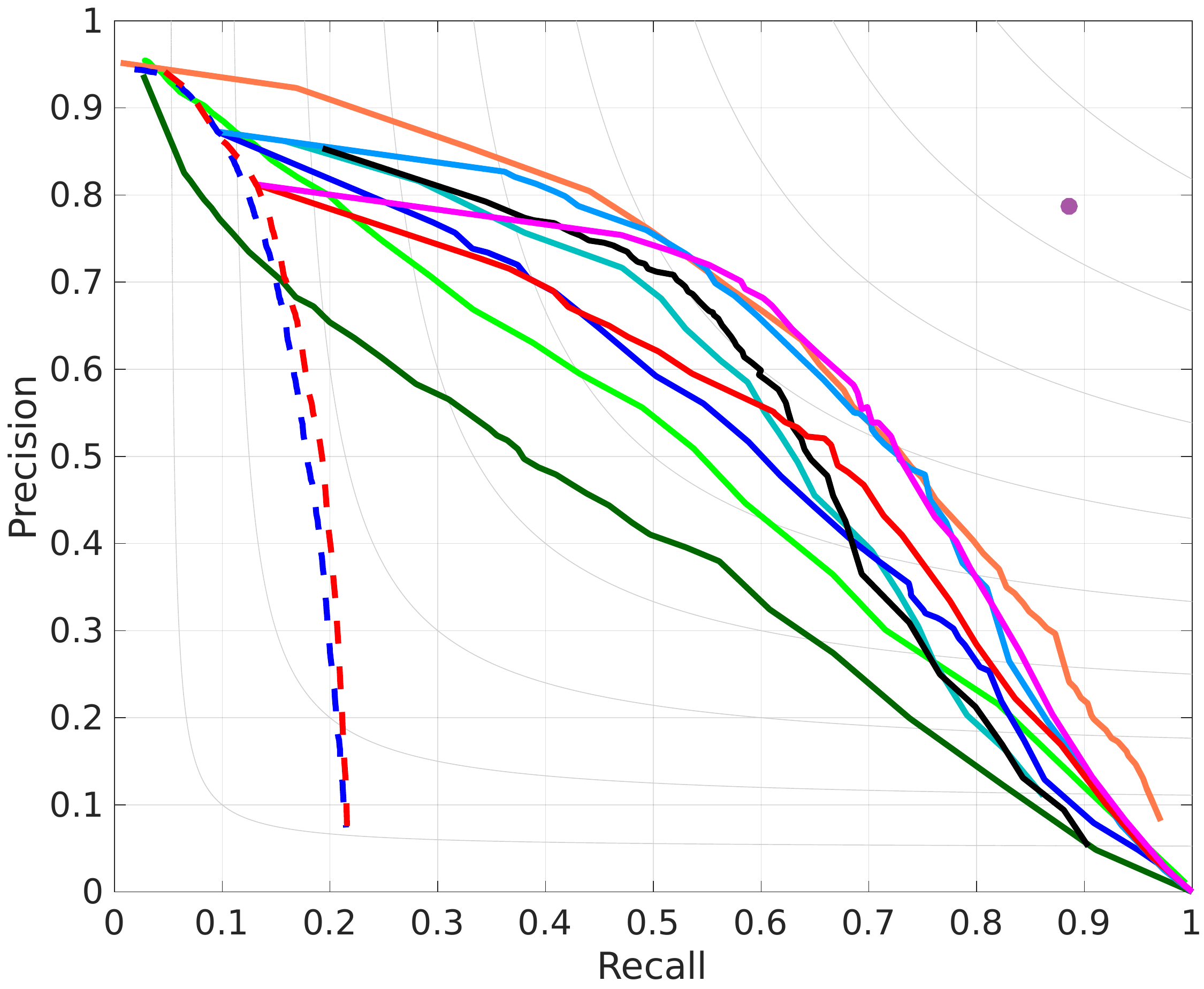}
\par\end{centering}
}\hfill{}\vspace{-0.5em}
\par\end{centering}
\caption{\label{fig:video-segmentation-test-set}Comparison of state-of-the-art
video segmentation algorithms with/without our improved superpixels,
on the test set of VSB100 \cite{Galasso13}. Dashed lines indicate
only frame-by-frame processing. See table \ref{tab:video-segmentation-test-set}
and \S\ref{subsec:Test-set-video-segmentation-results} for details.}
\vspace{-0.5em}
\end{figure*}
\begin{table}[!h]
\vspace{0em}
\setlength{\tabcolsep}{3.5pt} 
\centering{}{\scriptsize{}\hspace*{-1em}}%
\begin{tabular}{cccc|ccccc}
 & \multicolumn{3}{c|}{{\scriptsize{}BPR}} & \multicolumn{3}{c}{{\scriptsize{}VPR}} & {\scriptsize{}Length} & {\scriptsize{}NCL}\tabularnewline
\cline{2-7} 
{\scriptsize{}Algorithm} & {\scriptsize{}ODS} & {\scriptsize{}OSS} & {\scriptsize{}AP} & {\scriptsize{}ODS} & {\scriptsize{}OSS} & {\scriptsize{}AP} & {\scriptsize{}$\mu\left(\delta\right)$} & {\scriptsize{}$\mu$}\tabularnewline
\hline 
\hline 
{\scriptsize{}Human} & {\scriptsize{}0.81} & {\scriptsize{}0.81} & {\scriptsize{}0.67} & {\scriptsize{}0.83} & {\scriptsize{}0.83} & {\scriptsize{}0.70} & {\scriptsize{}83.2(40.0)} & {\scriptsize{}11.9}\tabularnewline
\hline 
{\scriptsize{}Grundmann et al. \cite{GrundmannKwatraHanEssaCVPR10}\hspace*{-1em}} & {\scriptsize{}0.47} & {\scriptsize{}0.54} & {\scriptsize{}0.41} & {\scriptsize{}0.52} & {\scriptsize{}0.55} & {\scriptsize{}0.52} & {\scriptsize{}87.7(34.0)} & {\scriptsize{}18.8}\tabularnewline
{\scriptsize{}Galasso et al.'12 \cite{GalassoCS12}} & {\scriptsize{}0.51} & {\scriptsize{}0.56} & {\scriptsize{}0.45} & {\scriptsize{}0.45} & {\scriptsize{}0.51} & {\scriptsize{}0.42} & {\scriptsize{}80.2(37.6)} & {\scriptsize{}8.0}\tabularnewline
{\scriptsize{}Yi and Pavlovic \cite{Yi2015Iccv}} & {\scriptsize{}0.63} & {\scriptsize{}0.67} & {\scriptsize{}0.60} & \textbf{\scriptsize{}\uline{0.64}} & \textbf{\scriptsize{}\uline{0.67}} & {\scriptsize{}{}0.65} & {\scriptsize{}35.83(38.9)} & {\scriptsize{}167.3}\tabularnewline
{\scriptsize{}Keuper and Brox \cite{Keuper16Arxiv}} & {\scriptsize{}0.56} & {\scriptsize{}0.63} & {\scriptsize{}0.56} & \textbf{\scriptsize{}\uline{0.64}} & {\scriptsize{}0.66} & \textbf{\scriptsize{}\uline{0.67}} & {\scriptsize{}1.1(0.7)} & {\scriptsize{}962.6}\tabularnewline
\hline 
{\scriptsize{}Segm. propagation \cite{Galasso13}\hspace*{-1em}} & {\scriptsize{}0.61} & {\scriptsize{}0.65} & {\scriptsize{}0.59} & {\scriptsize{}0.59} & {\scriptsize{}0.62} & {\scriptsize{}0.56} & {\scriptsize{}25.5(36.5)} & {\scriptsize{}258.1}\tabularnewline
{\scriptsize{}Our SPX + \cite{Galasso13}} & \textbf{\scriptsize{}0.64} & \textbf{\scriptsize{}0.69} & \textbf{\scriptsize{}\uline{0.67}} & \textbf{\scriptsize{}0.61} & \textbf{\scriptsize{}0.63} & \textbf{\scriptsize{}0.57} & {\scriptsize{}22.2(34.4)} & {\scriptsize{}216.8}\tabularnewline
\hline 
{\scriptsize{}Spectral graph reduction\cite{Galasso14}} & {\scriptsize{}0.62} & {\scriptsize{}0.66} & \textbf{\scriptsize{}0.54} & {\scriptsize{}0.55} & {\scriptsize{}0.59} & \textbf{\scriptsize{}0.55} & {\scriptsize{}61.3(40.9)} & {\scriptsize{}80.0}\tabularnewline
{\scriptsize{}Our SPX + \cite{Galasso14}} & \textbf{\scriptsize{}\uline{0.66}} & \textbf{\scriptsize{}0.68} & {\scriptsize{}0.51} & \textbf{\scriptsize{}0.58} & \textbf{\scriptsize{}0.61} & \textbf{\scriptsize{}0.55} & {\scriptsize{}70.4(40.2)} & {\scriptsize{}15.0}\tabularnewline
\hline 
{\scriptsize{}Graph construction \cite{KhorevaCVPR2015}} & {\scriptsize{}0.64} & \textbf{\scriptsize{}\uline{0.70}} & \textbf{\scriptsize{}0.61} & {\scriptsize{}0.63} & {\scriptsize{}0.66} & \textbf{\scriptsize{}0.63} & {\scriptsize{}83.4(35.3)} & {\scriptsize{}50.0}\tabularnewline
{\scriptsize{}Our SPX +\cite{KhorevaCVPR2015}} & \textbf{\scriptsize{}\uline{0.66}} & \textbf{\scriptsize{}\uline{0.70}} & {\scriptsize{}0.55} & \textbf{\scriptsize{}\uline{0.64}} & \textbf{\scriptsize{}\uline{0.67}} & {\scriptsize{}0.61} & {\scriptsize{}79.4(35.6)} & {\scriptsize{}50.0}\tabularnewline
\end{tabular}\vspace{0em}
\caption{\label{tab:video-segmentation-test-set}Comparison of state-of-the-art
video segmentation algorithms with our proposed method based on the
improved superpixels, on the test set of VSB100 \cite{Galasso13}.
The table shows BPR and VPR and length statistics (mean $\mu$, standard
deviation $\delta$, no. clusters NCL), see figure \ref{fig:video-segmentation-test-set}
and \S\ref{subsec:Test-set-video-segmentation-results} for details.}
\end{table}

\subsection{\label{subsec:Validation-set-video-segmentation-results}Validation
set results}

\textcolor{black}{We use two baseline methods (\cite{Galasso14,Galasso13},
see \S\ref{sec:Graph-based-video-segmentation}) to show the advantage
of using the proposed superpixels, although our approach is directly
applicable to any graph-based video segmentation technique. The baseline
methods originally employ the superpixels proposed by \cite{ArbelaezMaireFowlkesMalikPAMI11,GalassoCS12},
which use the boundary estimation $\mbox{gPb}_{\mathcal{I}}\negmedspace+\negmedspace\mbox{gPb}_{\mathcal{F}}$
to construct a segmentation. }

\textcolor{black}{For the baseline method of \cite{Galasso14} we
build a graph, where superpixels generated from the hierarchical image
segmentation based on the proposed boundary estimation are taken as
nodes. Following \cite{Galasso14} we select the hierarchy level of
image segmentation to extract superpixels (threshold over the ultrametric
contour map) by a grid search on the validation set. We aim for the
level which gives the best video segmentation performance, optimizing
for both BPR and VPR. }

Figure \ref{fig:video-segmentation-validation-set} presents results
on the validation set of VSB100. The dashed curves indicate frame-by-frame
segmentation and show (when touching the continuous curves) the chosen
level of hierarchy to extract superpixels. As it appears in the plots,
our superpixels help to improve video segmentation performance on
BPR and VPR for both baseline methods \cite{Galasso13,Galasso14}.
Figure \ref{fig:per-video-validation} shows the performance of video
segmentation with the proposed superpixels per video sequence. Our
method improves most on hard cases, where the performance of the original
approach was quite low, OSS less than 0.5. 

\subsection{\textcolor{black}{\label{subsec:Test-set-video-segmentation-results}Test
set results}}

\paragraph{\textcolor{black}{VSB100}}

Figure \ref{fig:video-segmentation-test-set} and Table \ref{tab:video-segmentation-test-set}
show the comparison of the baseline methods \cite{Galasso13,Galasso14}
with and without superpixels generated from the proposed boundaries,
and with state-of-the-art video segmentation algorithms on the test
set of VSB100. For extracting per-frame superpixels from the constructed
hierarchical segmentation we use the level selected on the validation
set. 

As shown in the plots and the table, the proposed method improves
the baselines considered. The segmentation propagation \cite{Galasso13}
method improves $\sim\negmedspace5$ percent points on the BPR metrics,
and $1\negmedspace\sim\negmedspace2$ points on the VPR metrics. This
supports that employing temporal cues helps to improve temporal consistency
across frames. Our superpixels also boosts the performance of the
approach from \cite{Galasso14}.\textcolor{black}{{} }
\begin{figure*}[t]
\begin{centering}
\hspace*{\fill}\includegraphics[bb=0bp 0bp 768bp 432bp,width=0.162\textwidth,height=0.07\textheight]{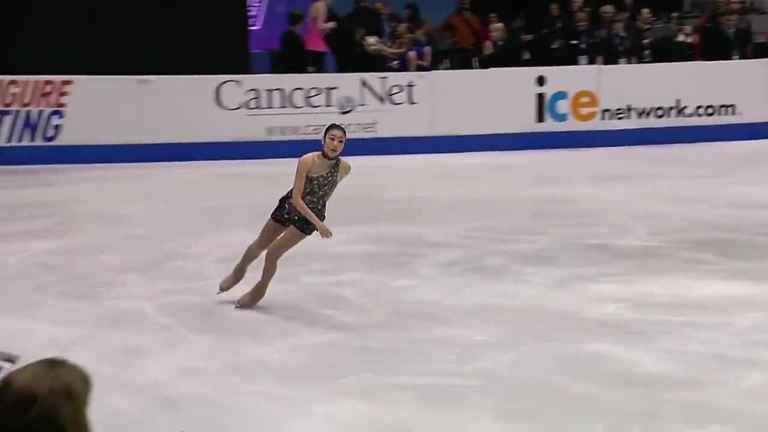}\,\includegraphics[bb=0bp 0bp 768bp 432bp,width=0.162\textwidth,height=0.07\textheight]{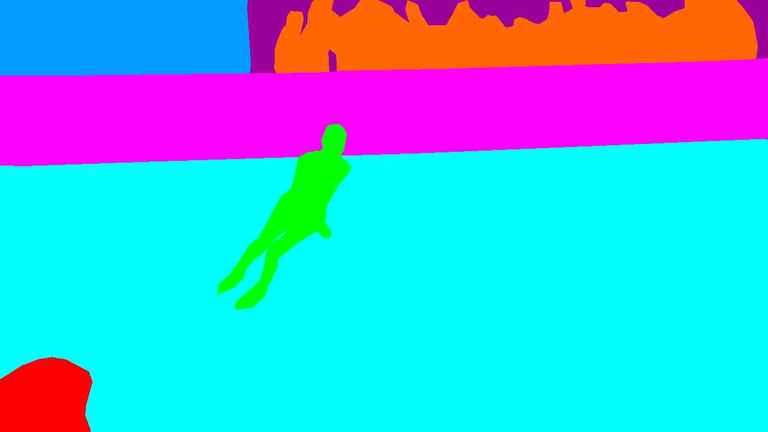}\,\includegraphics[width=0.162\textwidth,height=0.07\textheight]{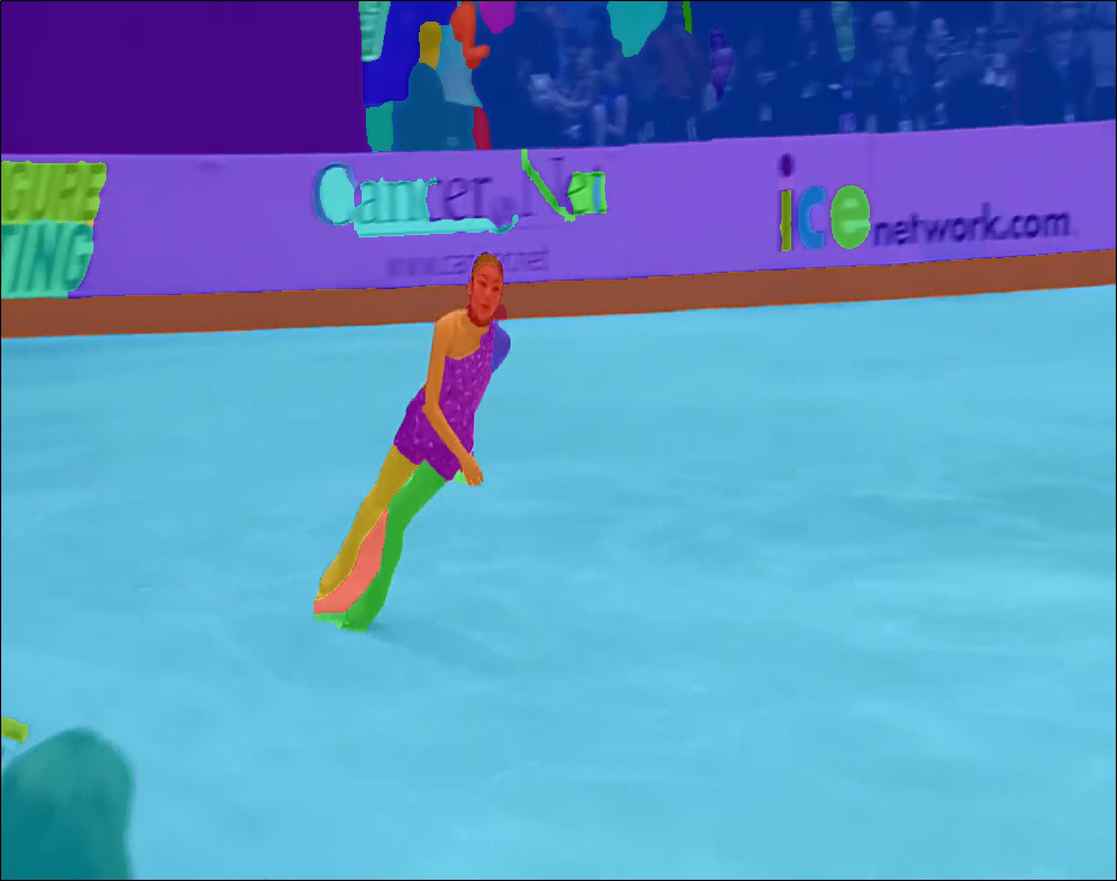}\,\includegraphics[width=0.162\textwidth,height=0.07\textheight]{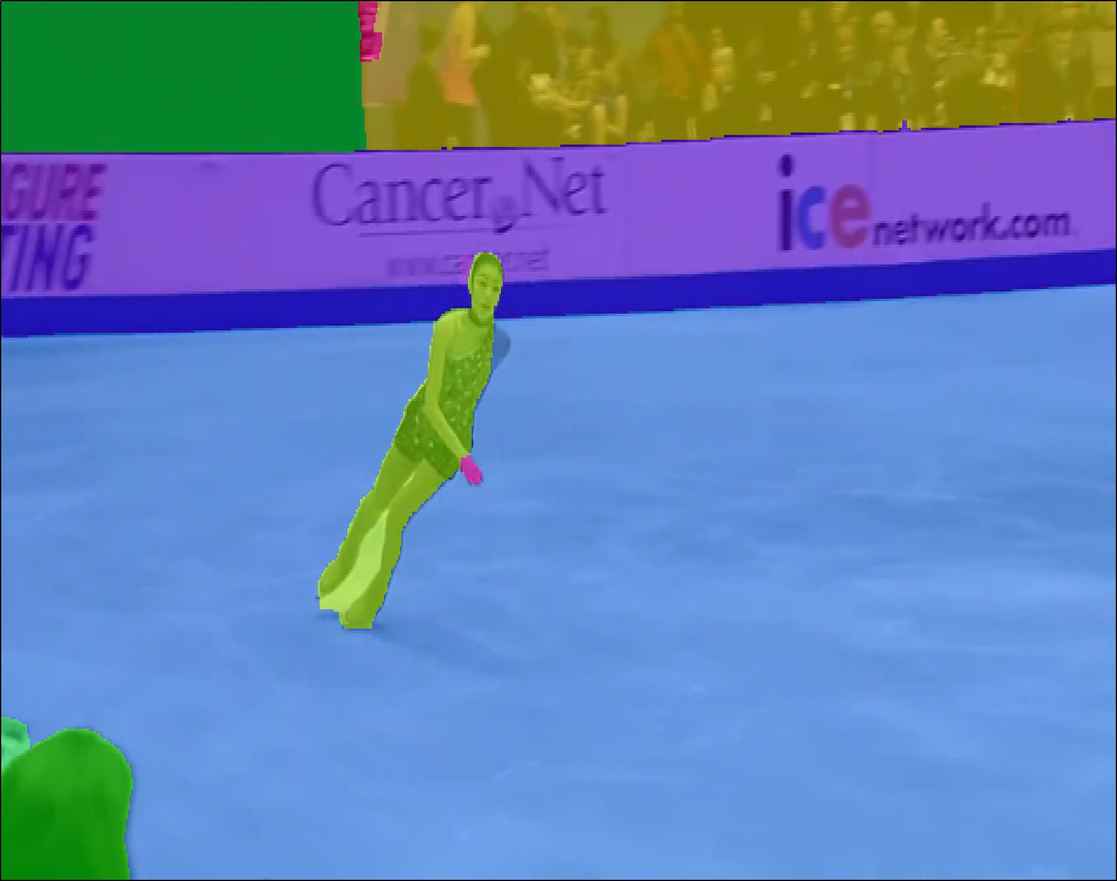}\,\includegraphics[width=0.162\textwidth,height=0.07\textheight]{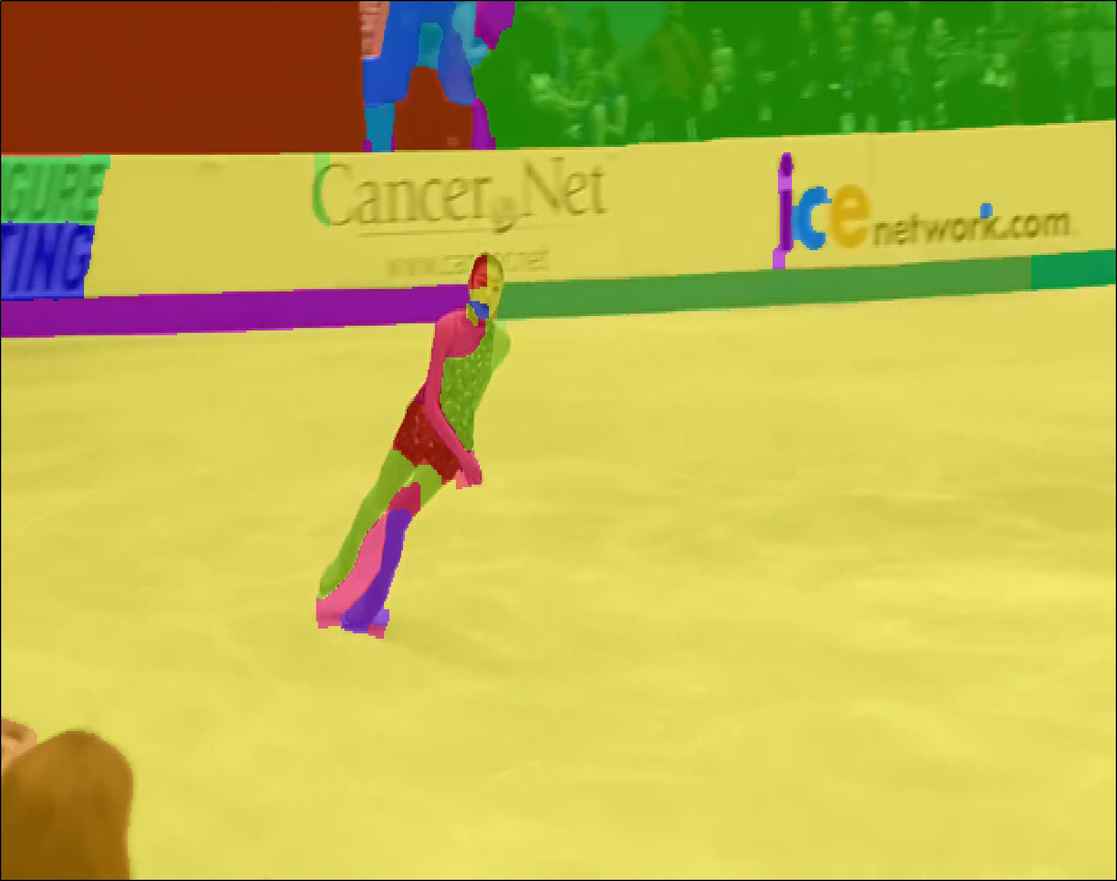}\,\includegraphics[width=0.162\textwidth,height=0.07\textheight]{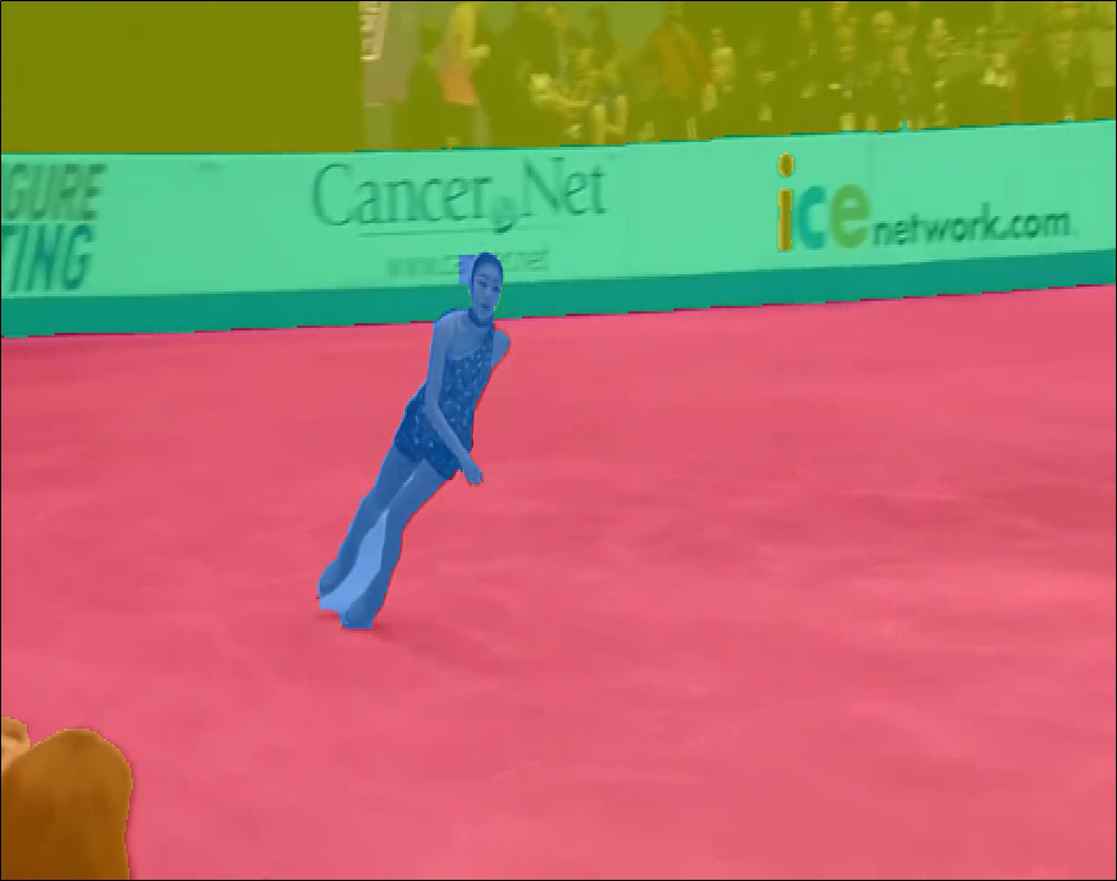}\hspace*{\fill}
\par\end{centering}
\begin{centering}
\hspace*{\fill}\includegraphics[bb=0bp 0bp 768bp 432bp,width=0.163\textwidth]{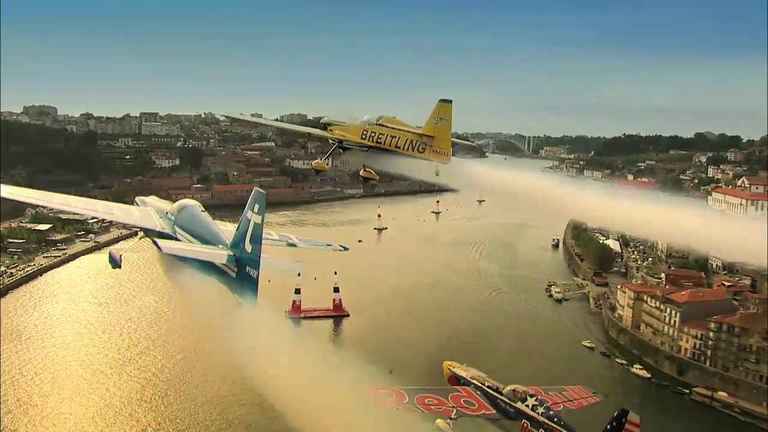}\,\includegraphics[bb=0bp 0bp 768bp 432bp,width=0.162\textwidth]{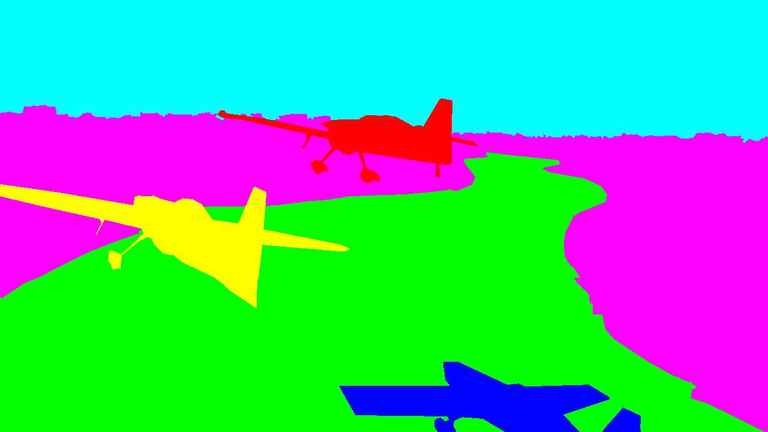}\,\includegraphics[width=0.162\textwidth]{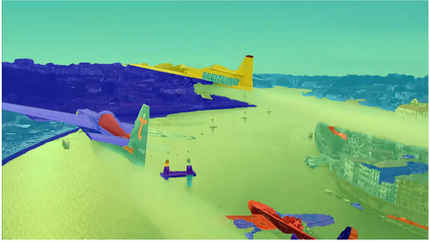}\,\includegraphics[width=0.162\textwidth]{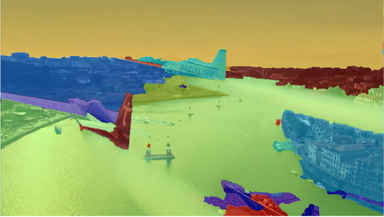}\,\includegraphics[width=0.162\textwidth]{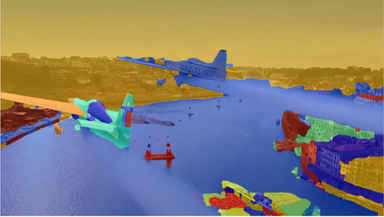}\,\includegraphics[width=0.162\textwidth]{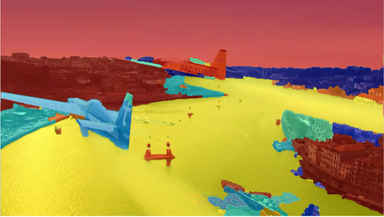}\hspace*{\fill}
\par\end{centering}
\begin{centering}
\hspace*{\fill}\includegraphics[width=0.162\textwidth,height=0.07\textheight]{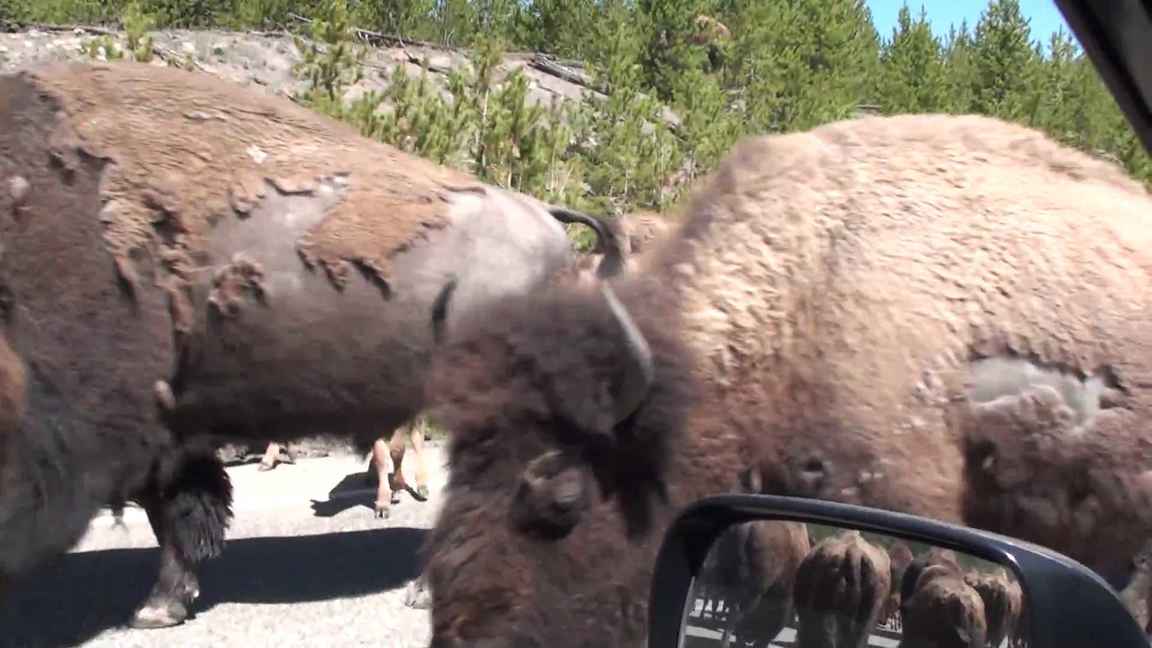}\,\includegraphics[bb=0bp 0bp 1152bp 648bp,clip,width=0.162\textwidth,height=0.07\textheight]{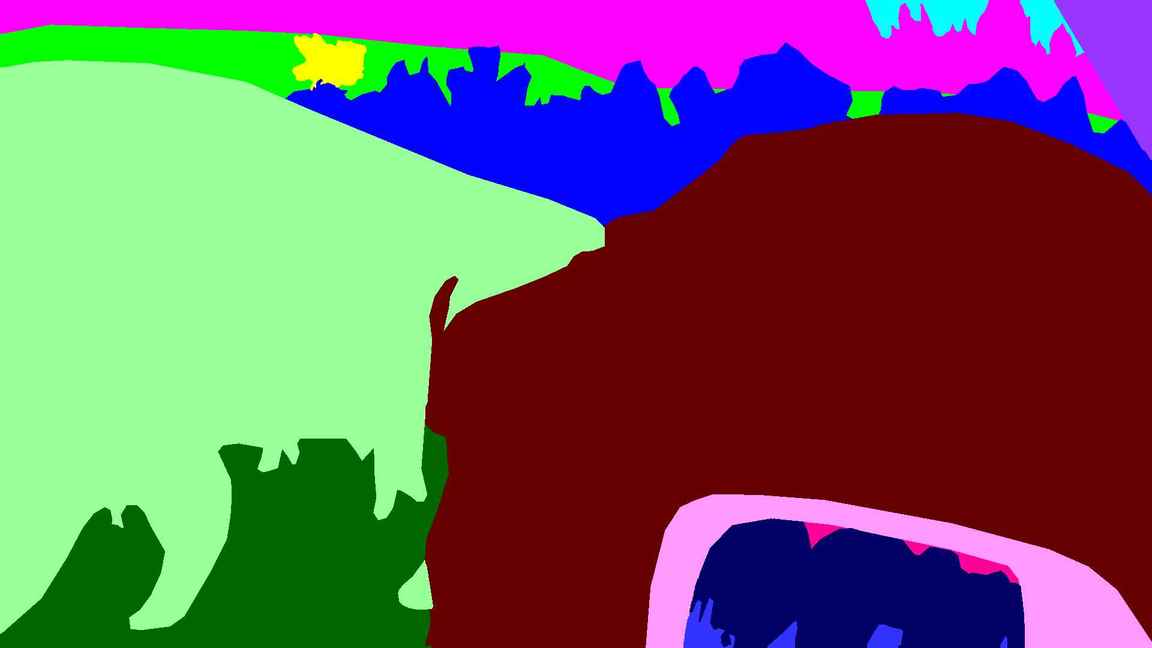}\,\includegraphics[width=0.162\textwidth,height=0.07\textheight]{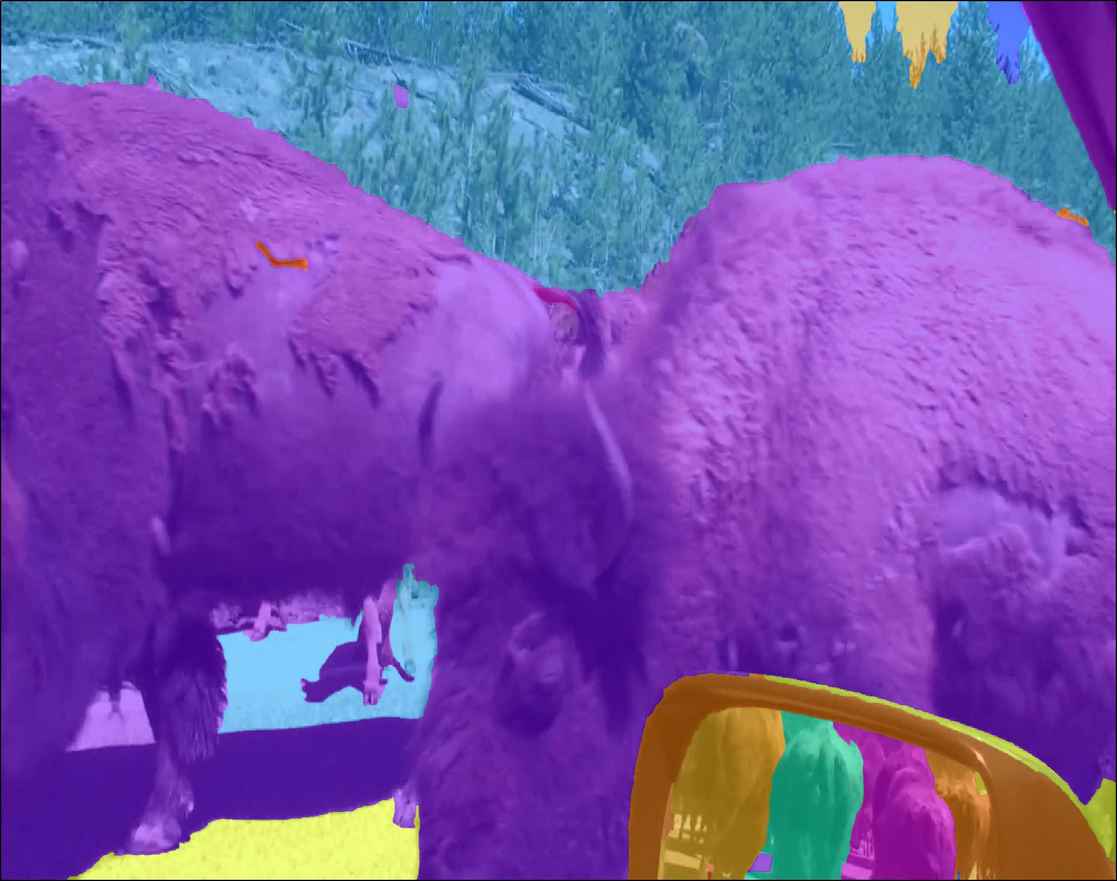}\,\includegraphics[clip,width=0.162\textwidth,height=0.07\textheight]{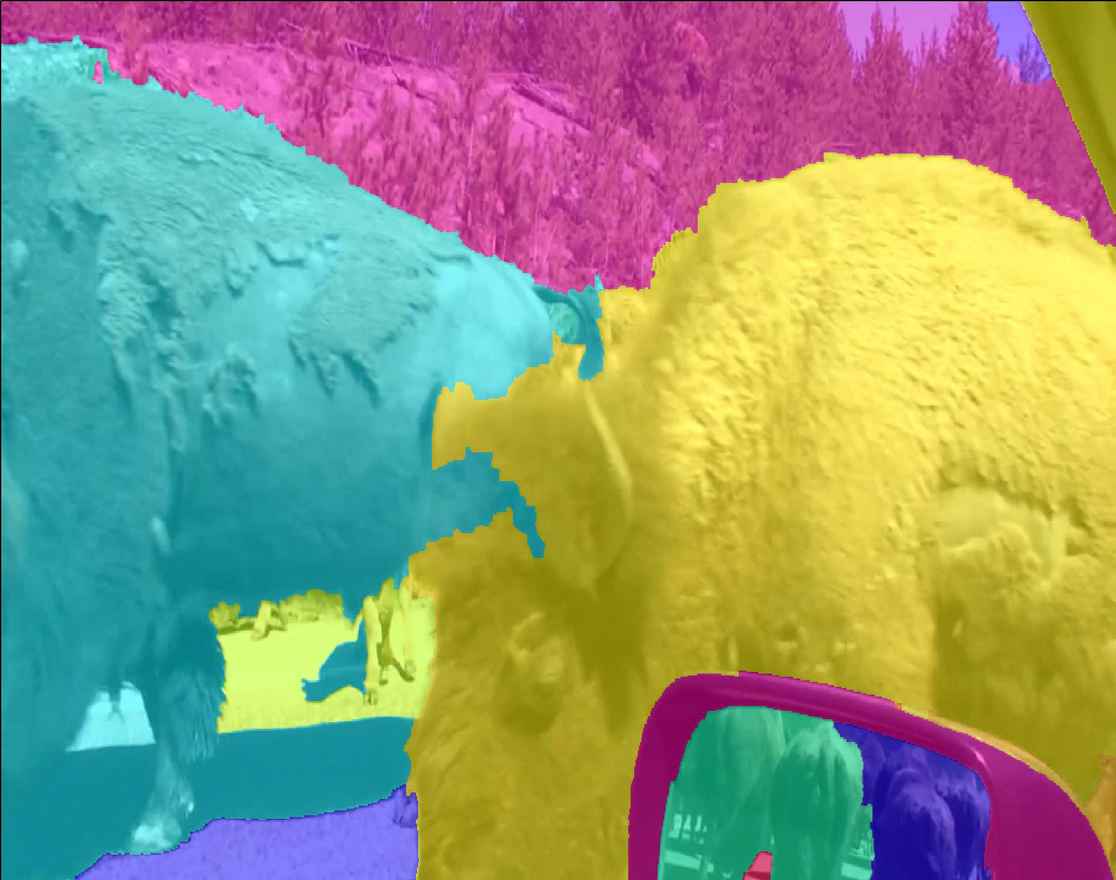}\,\includegraphics[width=0.162\textwidth,height=0.07\textheight]{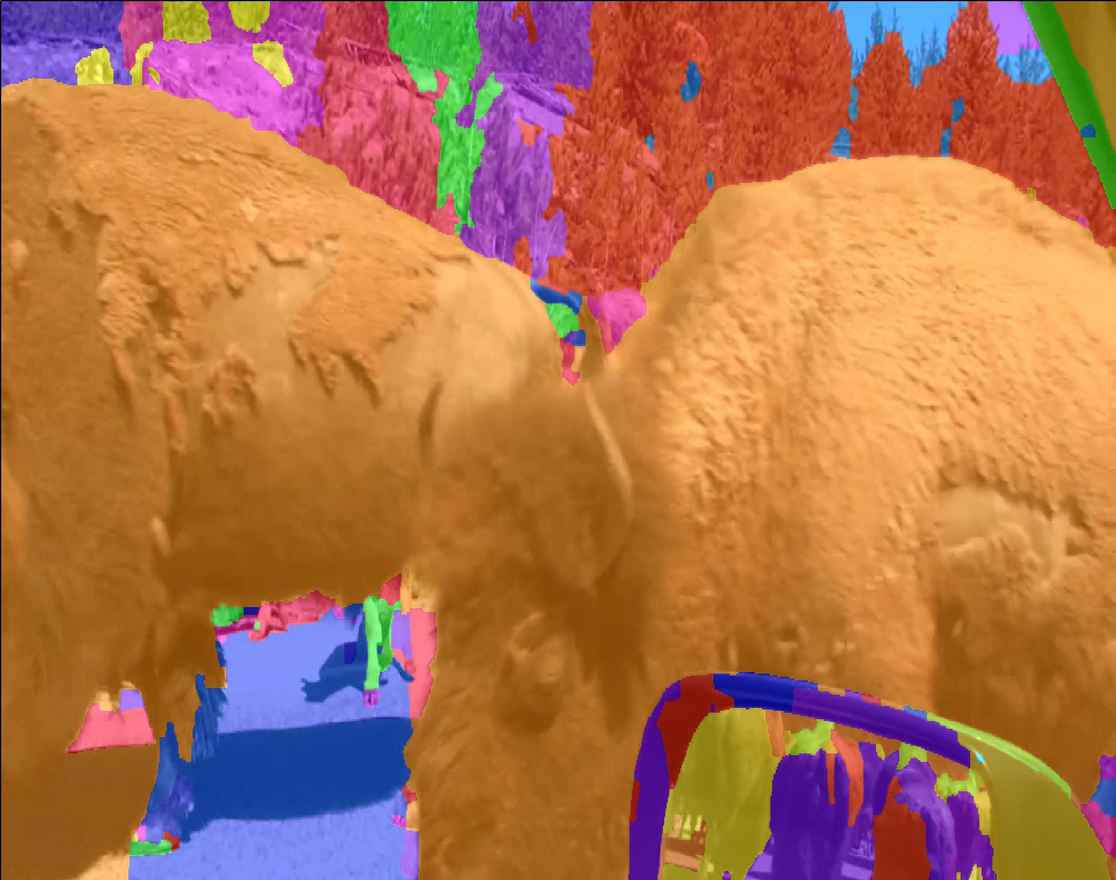}\,\includegraphics[width=0.162\textwidth,height=0.07\textheight]{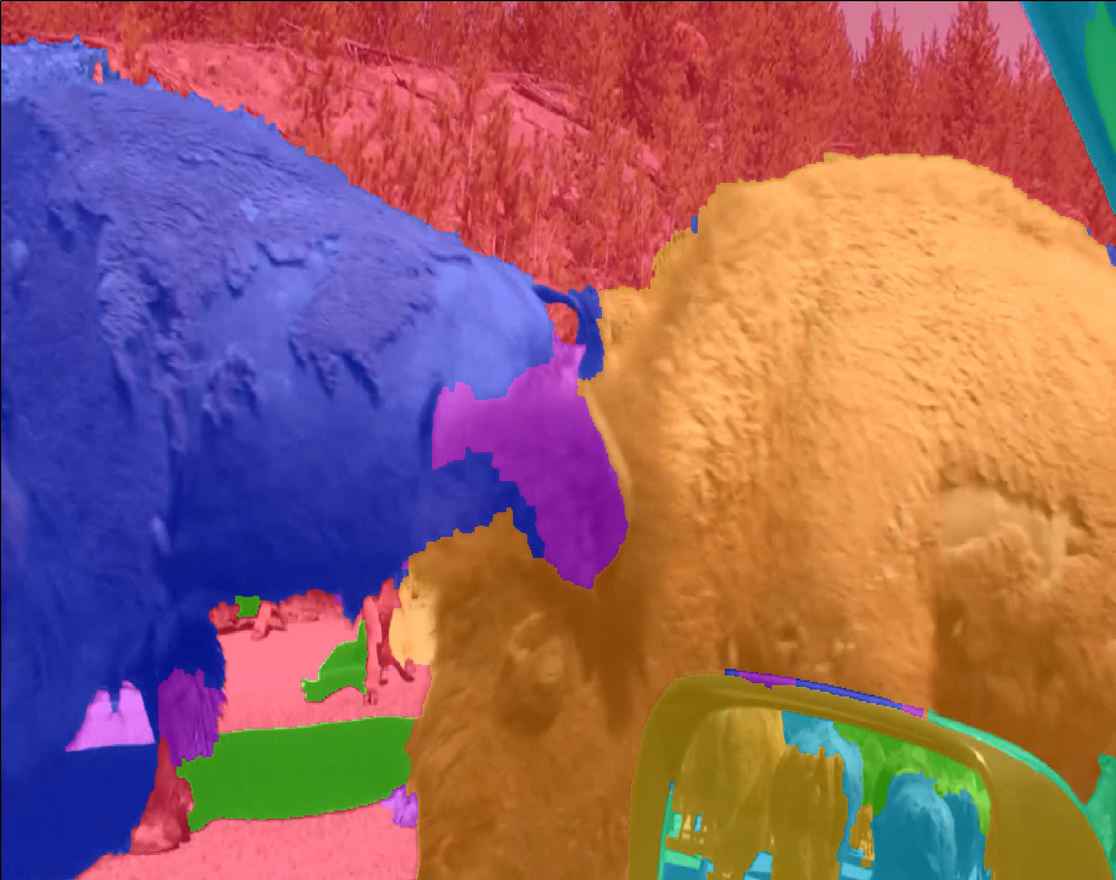}\hspace*{\fill}
\par\end{centering}
\begin{centering}
\hspace*{\fill}\includegraphics[bb=0bp 0bp 768bp 432bp,width=0.163\textwidth]{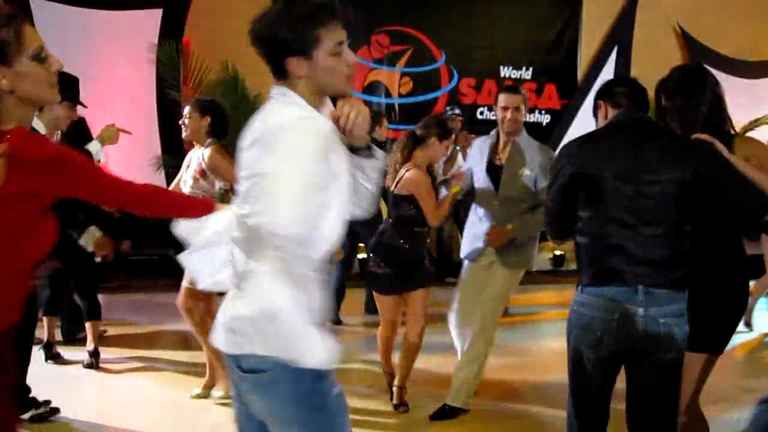}\,\includegraphics[bb=0bp 0bp 768bp 432bp,width=0.162\textwidth]{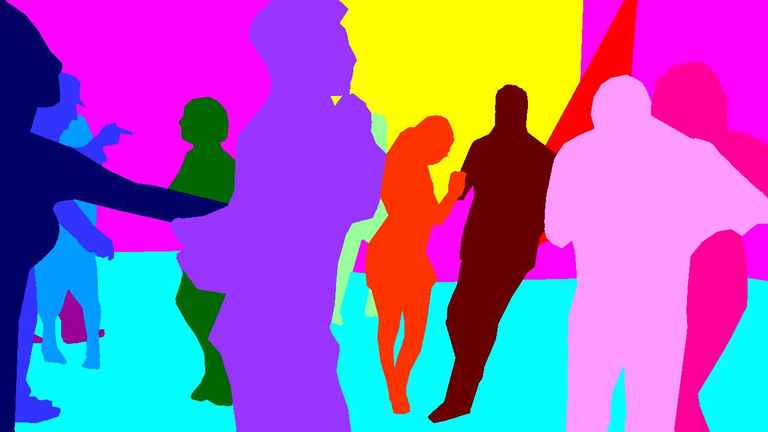}\,\includegraphics[width=0.162\textwidth]{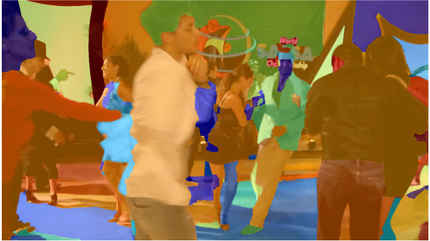}\,\includegraphics[width=0.162\textwidth]{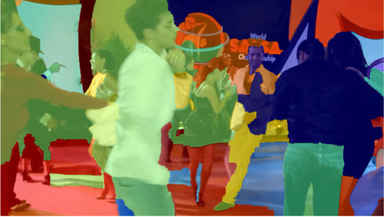}\,\includegraphics[width=0.162\textwidth]{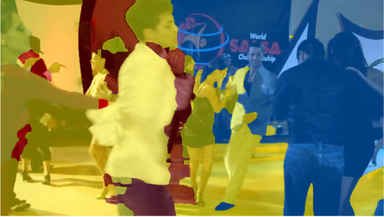}\,\includegraphics[width=0.162\textwidth]{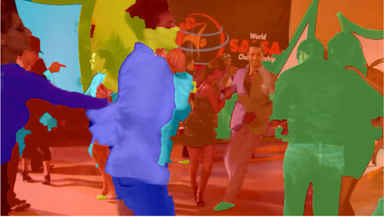}\hspace*{\fill}
\par\end{centering}
\hspace*{\fill}%
\begin{minipage}[t]{0.163\textwidth}%
\begin{center}
Video
\par\end{center}%
\end{minipage}\,%
\begin{minipage}[t]{0.163\textwidth}%
\begin{center}
GT
\par\end{center}%
\end{minipage}\,%
\begin{minipage}[t]{0.163\textwidth}%
\begin{center}
\cite{Galasso13}
\par\end{center}%
\end{minipage}\,%
\begin{minipage}[t]{0.163\textwidth}%
\begin{center}
Our SPX+\cite{Galasso13}
\par\end{center}%
\end{minipage}\,%
\begin{minipage}[t]{0.163\textwidth}%
\begin{center}
\cite{Galasso14}
\par\end{center}%
\end{minipage}\,%
\begin{minipage}[t]{0.163\textwidth}%
\begin{center}
Our SPX+\cite{Galasso14}
\par\end{center}%
\end{minipage}\hspace*{\fill}
\begin{centering}
\vspace{-0.5em}
\par\end{centering}
\caption{\label{fig:comparative-examples}Comparison of video segmentation
results of \cite{Galasso13,Galasso14} with our proposed superpixels
to one human ground truth. The last row shows a failure case for all
methods.}
\vspace{-1em}
\end{figure*}
\begin{figure}[t]
\begin{centering}
\vspace{-1em}
\hspace*{\fill}\subfloat[\label{fig:bpr-test-set-BSDM}BPR on BMDS test set]{\begin{centering}
\includegraphics[bb=0bp 0bp 597bp 490bp,width=0.47\columnwidth]{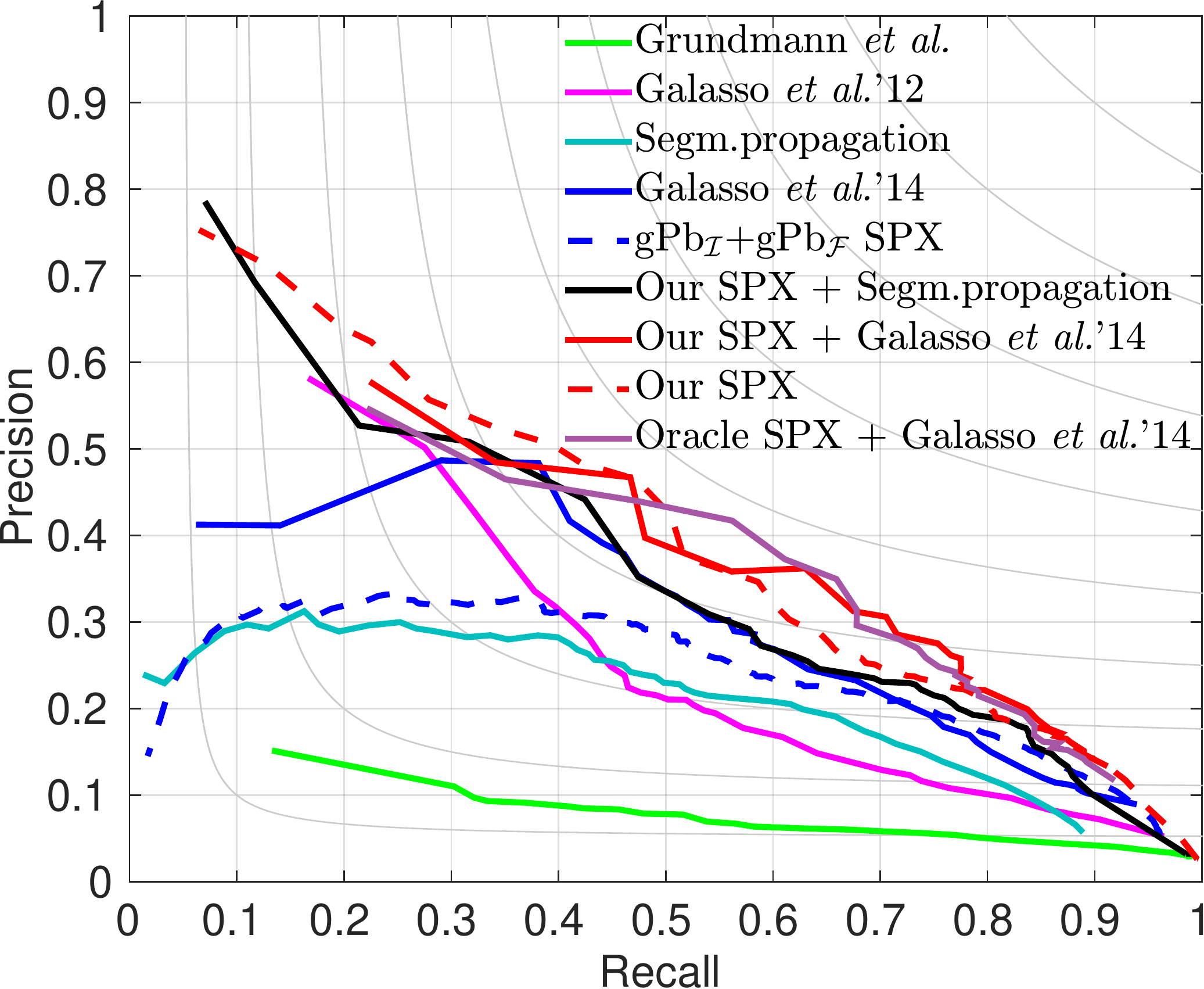}
\par\end{centering}
}\hspace*{\fill}\subfloat[\label{fig:vpr-test-set-BSDM}VPR on BMDS test set]{\begin{centering}
\includegraphics[bb=0bp 0bp 599bp 490bp,width=0.47\columnwidth]{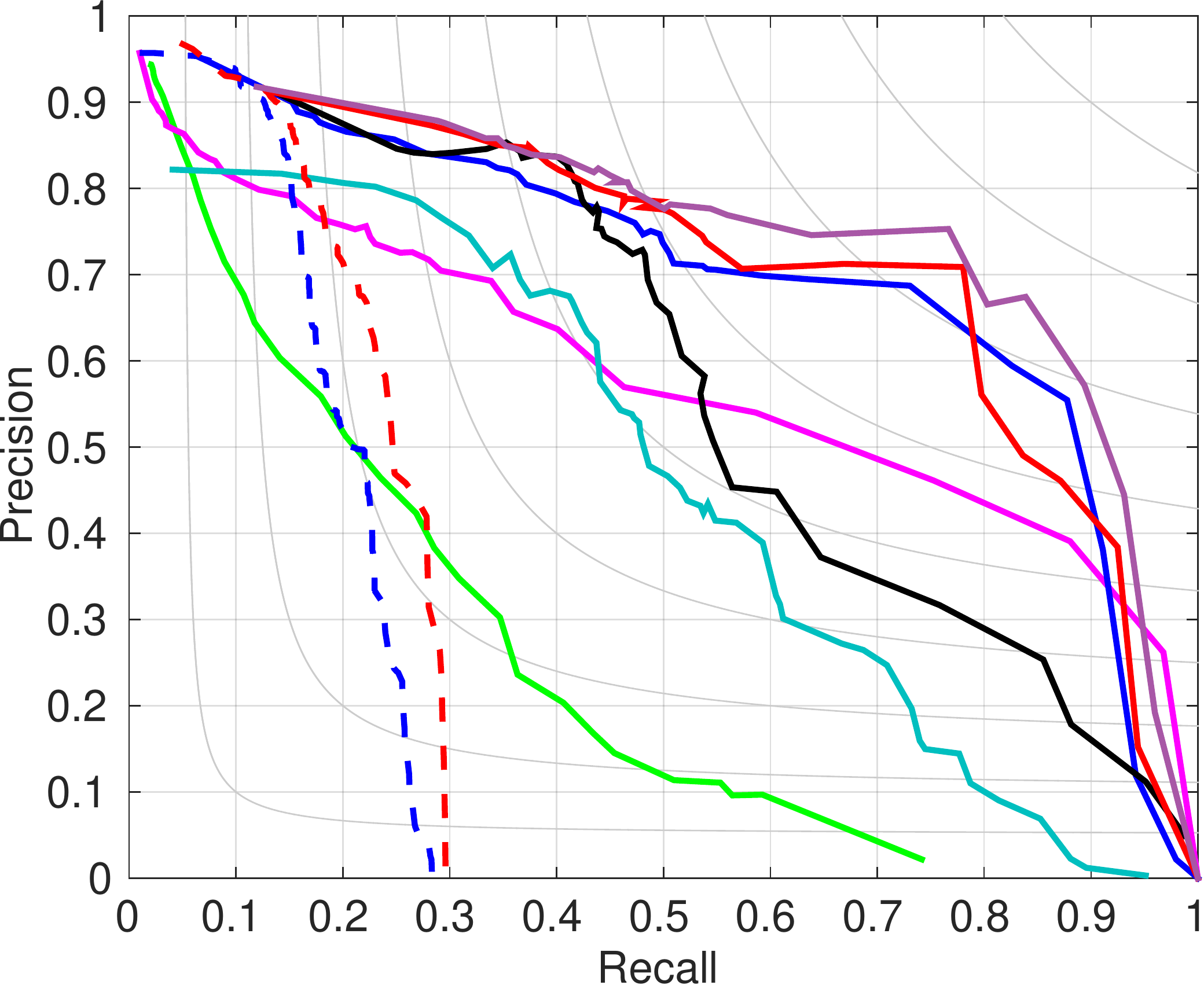}
\par\end{centering}
}\hspace*{\fill}\vspace{-0.5em}
\par\end{centering}
\caption{\label{fig:video-segmentation-test-set-bsdm}Comparison of state-of-the-art
video segmentation algorithms with the proposed superpixels, on BMDS
\cite{BroxMalikECCV10}. Dashed lines indicate only frame-by-frame
processing (see \S\ref{subsec:Test-set-video-segmentation-results}
for details).}
\vspace{-0.5em}
\end{figure}

Employing our method for graph-based video segmentation also benefits
computational load, since it depends on the number of nodes in the
graph (number of generated superpixels). On average the number of
nodes is reduced by a factor of $2.6$, 120 superpixels per frame
versus 310 in \cite{Galasso14}. \textcolor{black}{This leads to $\sim45\%$
reduction in runtime and memory usage for video segmentation.}

\textcolor{black}{Given the videos and their optical flow, the superpixel
computation takes 90\% of the total time and video segmentation only
10\% (for both \cite{Galasso14} and our SPX+\cite{Galasso14}). Our
superpixels are computed $20\%$ faster than $\mbox{gPb}_{\mathcal{I}}\negmedspace+\negmedspace\mbox{gPb}_{\mathcal{F}}$
(the bulk of the time is spent in $\mbox{OP}\left(\cdot\right)$).
The overall time of our approach is $20\%$ faster than \cite{Galasso14}.
}

Qualitative results are shown in Figure \ref{fig:comparative-examples}.
Superpixels generated from the proposed boundaries\textcolor{red}{{}
}allow the baseline methods \cite{Galasso13,Galasso14} to better
distinguish visual objects and to limit label leakage due to inherent
temporal smoothness of the boundaries. Qualitatively the proposed
superpixels improve video segmentation on easy (e.g. first row of
Figure \ref{fig:comparative-examples}) as well as hard cases (e.g.
second row of Figure \ref{fig:comparative-examples}).

\textcolor{black}{As our approach is directly applicable to any graph-based
video segmentation technique} we additionaly evaluated our superpixels
with the classifier-based graph construction method of \cite{KhorevaCVPR2015}.
The method learns the topology and edge weights of the graph using
features of superpixels extracted from per-frame segmentations. We
employed this approach without re-training the classifiers on the
proposed superpixels. Using our superpixels alows to achieve on par
performance (see Figure \ref{fig:video-segmentation-test-set} and
Table \ref{tab:video-segmentation-test-set}) while significantly
reducing the runtime and memory load (\textcolor{black}{$\sim45\%$)}.
\textcolor{black}{Superpixels based on per-frame boundary estimation
are also employed in }\cite{Yi2015Iccv}. However\textcolor{black}{{}
we could not evaluate its performance }with \textcolor{black}{our
superpixels as} the code is not available under open source.

\paragraph{BMDS}

Further we evaluate the proposed method on BMDS \cite{BroxMalikECCV10}
to show the generalization of our superpixels across datasets. We
use the same model trained on VSB100 for generating superpixels and
the hierarchical level of boundary map as validated by a grid search
on the training set of BMDS. The results are presented in Figure \ref{fig:video-segmentation-test-set-bsdm}.
Our boundaries based superpixels boost the performance of the baseline
methods \cite{Galasso13,Galasso14}, particularly for the BPR metric
(up to 4-12\%). 

\paragraph{Oracle}

Additionally we set up the oracle case for the baseline \cite{Galasso14}
(purple curve in Figure \ref{fig:video-segmentation-test-set-bsdm})
by choosing the hierarchical level to extract superpixels from the
boundary map for each video sequence individually based on its performance
(we considered OSS measures for BPR and VPR of each video). The oracle
result indicates that the used fixed hierarchical level is quite close
to an ideal video-per-video selection.

\vspace{0em}

\section{\label{sec:Conclusion}Conclusion}

The presented experiments have shown that boundary based superpixels,
extracted via hierarchical image segmentation, are a better starting
point for graph-based video segmentation than classical superpixels.
However, the segmentation quality depends directly on the quality
of the initial boundary estimates.

Over the state-of-the-art methods such as $\mbox{\ensuremath{\mbox{SE}_{\mathcal{I}}}}$
\cite{Dollar2015PAMI}, our results show that we can significantly
improve boundary estimates when using cues from object proposals,
globalization, and by merging with optical flow cues. When using superpixels
built over these improved boundaries, we observe consistent improvement
over two different video segmentation methods \cite{Galasso13,Galasso14}
and two different datasets (VSB100, BMDS). The results analysis indicates
that we improve most in the cases where baseline methods degrade.

\textcolor{black}{}For future work we are encouraged by the promising
results of object proposals. We believe that there is room for further
improvement by integrating more semantic notions of objects into video
segmentation.

\bibliographystyle{ieee}
\bibliography{2016_cvpr_video_segmentation}

\end{document}